\documentclass[letterpaper, 10 pt, conference]{ieeeconf} 
\IEEEoverridecommandlockouts  
\usepackage[T1]{fontenc}
\usepackage{amsmath,amsfonts}
\usepackage{algorithmic}
\usepackage{textcomp}
\usepackage{stfloats}
\usepackage{url}
\usepackage{verbatim}
\usepackage{graphicx}
\usepackage{cite}
\usepackage{bm}
\usepackage{color}
\makeatletter
\let\NAT@parse\undefined
\makeatother
\usepackage[colorlinks=true,
linkcolor=blue,
anchorcolor=blue,
citecolor=blue]{hyperref}
\usepackage{flushend}
\usepackage{amsmath}
\usepackage{subcaption}
\usepackage{booktabs}
\usepackage{float}
\usepackage{afterpage}
\begin{document}
\title{Multi-Agent Target Assignment and Path Finding for Intelligent Warehouse: A Cooperative Multi-Agent Deep Reinforcement Learning Perspective}

\author{
    Qi Liu$^{1}$, Jianqi Gao$^{1}$, Dongjie Zhu$^{2}$, Zhongjian Qiao$^{3}$, Pengbin Chen$^{1}$, Jingxiang Guo$^{1}$, Yanjie Li\textsuperscript{1*}
    \thanks{This work was supported by the National Natural Science Foundation of China [61977019, U1813206] and Shenzhen Fundamental Research [JCYJ20220818102415033, JSGG20201103093802006, KJZD20230923114222045]. \textit{(Corresponding author: Yanjie Li, autolyj@hit.edu.cn)}}
    \thanks{${1}$\ Qi Liu, Jianqi Gao, Pengbin Chen, Jingxiang Guo, and Yanjie Li are with the Guangdong Key Laboratory of Intelligent Morphing Mechanisms and Adaptive Robotics and the School of Mechanical Engineering and Automation, the Harbin Institute of Technology Shenzhen, 518055, China.}
    \thanks{${2}$\ Dongjie Zhu is with the Department of Automation, Tsinghua University, 100084, China.}
    \thanks{${3}$\ Zhongjian Qiao is with the Tsinghua Shenzhen International Graduate School, Tsinghua University, 518055, China.}
}

% The paper headers
% \markboth{
% 	IEEE Robotics and Automation Letters. Pre-print version. July, 2024
% 	}%
% {Shell \MakeLowercase{\textit{et al.}}: A Sample Article Using IEEEtran.cls for IEEE Journals}

%\IEEEpubid{0000--0000/00\$00.00~\copyright~2022 IEEE}
% Remember, if you use this, you must call \IEEEpubidadjcol in the second
% column for its text to clear the IEEEpubid mark.

\maketitle
\begin{abstract}
Multi-agent target assignment and path planning (TAPF) are two key problems in intelligent warehouses. However, most literature only addresses one of these two problems separately. This paper proposes a method to simultaneously solve target assignment and path planning from the cooperative multi-agent deep reinforcement learning (RL) perspective. This paper models the TAPF problem for intelligent warehouses to cooperative multi-agent deep RL and simultaneously addresses task assignment and path finding based on multi-agent deep RL. Furthermore, previous literature rarely considers the physical dynamics of agents. In this paper, the physical dynamics of the agents are considered. This is more in line with actual robot movement. Experimental results show that the proposed method performs well in various task settings, which means that the target assignment is solved reasonably well and the planned path is almost the shortest. Moreover, the proposed method is more time-efficient than baselines.
\end{abstract}

% \begin{IEEEkeywords}
% Robot control, quadruped robot locomotion, reinforcement learning (RL), multi-agent RL.
% \end{IEEEkeywords}

\section{Introduction}
\label{Section: Introduction}

With the rapid development of logistics-related industries, new opportunities, and challenges have been proposed for intelligent warehousing \cite{adhau2012multi,ayari2019acd3gpso}. Traditional warehouse technologies use conveyor belts and other equipment to complete material handling. They are inflexible and not easily extensible. Intelligent warehousing utilizes a multi-agent system to deliver goods to designated locations, greatly improving efficiency. The intelligent warehouse system is divided mainly into the order fulfillment system, such as the kiva system \cite{wurman2008coordinating} and the logistics sorting center \cite{wan2018lifelong}. In order fulfillment systems, mobile agents move inventory pods to inventory stations. Then, workers take goods from inventory pods, and agents move the inventory pods to their original locations. In logistics sorting centers, agents take goods from loading stations and deliver them to chutes in the center of the warehouse. The approach proposed in this study is based on the logistics sorting center.

Task assignment and path finding (TAPF) are two important processes in intelligent warehousing. The system first assigns a specified task to each agent according to the order requirements. Then, the agent transports the goods from the origin to the destination and ensures that the path does not conflict with other agents. The task assignment and path planning problem is typically NP-hard \cite{ma2016multi}, which has a large search space. Thus, directly solving this problem is difficult. In general, two steps are required to solve the TAPF problem. The first step is multi-agent task assignment (MATA), assigning tasks to agents without considering possible path conflicts between agents. The second step is path planning for all agents by using multi-agent path finding (MAPF) algorithms. Numerous studies in the literature study MATA and MAPF; we describe these methods in detail in the next section. Although solving TAPF separately reduces the difficulty of the total problem, this type of method ignores the mutual influence between task assignment and path planning. Reasonable task assignments can reduce warehouse agents' path length, improve operation efficiency, and help avoid path conflicts between agents. Therefore, it is necessary to solve the TAPF problems together. In the TAPF problem, we assume that each idle agent and inventory pod are homogeneous. Thus, we can assign any task to any agent.

Recently, deep reinforcement learning (RL) has received great attention due to its remarkable successes in widely challenging domains, such as robotic control \cite{silver2014deterministic,10476692,10508809}, multi-agent problems \cite{rashid2018qmix}, and video games \cite{10466624}. In this study, we introduce cooperative multi-agent deep RL to solve the TAPF problem and address it simultaneously.

The main contributions of this paper can be summarized as follows:
\begin{itemize}
    \item This paper proposes a method to simultaneously solve target assignment and path planning from the cooperative multi-agent deep RL perspective. This paper models the TAPF problem for intelligent warehouses to cooperative multi-agent deep RL and simultaneously solves task assignment and path finding based on the multi-agent deep RL algorithm.
    \item Furthermore, in this paper, the physical dynamics of the agents are considered in the path planning phase. This is more in line with actual robot movement.  
    \item Experimental results show that the proposed method performs well in various task settings. This means the target assignment is solved reasonably, and the planned path is almost the shortest. Furthermore, the proposed method is more time-efficient than the baselines.
\end{itemize}

% The rest of this study is organized as follows. Section \ref{Related Work} discusses the related work of TAPF in an intelligent warehouse. Section \ref{Preliminary} introduces the preliminaries of the Markov decision process, RL, the cooperative multi-agent deep RL, and the definition of MAPF and TAPF problems. Section \ref{Multi-Agent Target Assignment and Path Finding for the Intelligent Warehouse: A Multi-agent Deep Reinforcement Learning Perspective} describes our method. Section \ref{Experiments} provides experimental results to verify the improved efficiency of the proposed method. Section \ref{Conclusions} presents conclusions and future work.

\section{Related Work}
\label{Related Work}

\subsection{Multi-Agent Task Assignment}
\label{Multi-Agent Task Assignment}
Through multi-agent task assignment algorithms, we can maximize the utilization of warehouse resources \cite{gerkey2004formal}. Currently, multi-agent task assignment algorithms in intelligent warehouses can be classed into centralized and distributed classes according to the management mode. In the centralized task assignment class, a central control system is set up, which is responsible for task assignment, assigning tasks assigned to agents for execution \cite{liu2012centralized}. Classical centralized task assignment algorithms include: Hungarian algorithm \cite{glover1990tabu}, tabu search algorithm \cite{kuhn1955hungarian}, genetic algorithm \cite{liu2012centralized}, etc. In the distributed assignment class, each agent in the warehouse plans its own task sequence according to tasks and environmental information \cite{giordani2013distributed}. This class method effectively reduces the load of the central control system and is more flexible and adaptable, but it may not find the global optimal solution \cite{best2019dec}. Furthermore, the distributed assignment class mainly includes learning-based methods \cite{bernstein2002complexity} and market auction methods \cite{zavlanos2008distributed}.

\subsection{Multi-Agent Path Finding}
\label{Multi-Agent Path Finding}
In recent years, MAPF has become a hot research direction in computer science and robotics. Classical MAPF algorithms can be divided into optimal and sub-optimal types according to whether the solution results meet the optimality. The optimal typical MAPF algorithms include A*-based \cite{wagner2015subdimensional}, conflict-based search \cite{sharon2015conflict}, increasing cost tree search based \cite{sharon2013increasing}, and compilation-based \cite{nguyen2019generalized} methods. Sub-optimal classical MAPF algorithms include search-based \cite{silver2005cooperative}, rule-based \cite{surynek2009novel}, compilation-based \cite{surynek2017modifying}, bounded conflict-based search \cite{barer2014suboptimal}, and genetic algorithm-based \cite{peihuang2009path} methods. However, traditional MAPF algorithms have poor real-time performance. Therefore, numerous researchers are beginning to study MAPF based on deep RL \cite{sutton2018reinforcement} to solve this problem. PRIMAL \cite{sartoretti2019primal} is a typical MAPF algorithm based on deep RL. However, PRIMAL still solves the multi-agent problem by single-agent deep RL method \cite{sartoretti2019primal}. In this study, we propose a method to simultaneously address target assignment and path planning from the cooperative multi-agent deep RL perspective. Furthermore, in the simulation environment, the above-mentioned various MAPF algorithms rarely consider the physical dynamics of agents. In this study, the physical dynamics factors of the agent are considered.

\subsection{Cooperative Multi-Agent Deep RL}
\label{Multi-Agent Reinforcement Learning}
Cooperative multi-agent RL (MARL) deals with systems consisting of several agents that interact in a common environment. Cooperative multi-agent RL aims to learn a policy to make all agents work together to complete a common goal. Multi-agent deep RL has received great attention due to its capability to allow agents to learn to make decisions in multi-agent environments through interactions. Independent Q-learning (IQL) \cite{tampuu2017multiagent} solves the multi-agent RL problem by decomposing it into a collection of simultaneous single-agent RL problems that share the same environment. However, IQL cannot solve the nonstationarity \cite{hernandez2017survey} problem caused by the changing policies of other agents. Value-Decomposition Networks (VDN) \cite{sunehag2018value} proposes to learn to linearly decompose a team value function into a series of per-agent value functions. After that, QMIX \cite{rashid2018qmix} utilizes a mixing network architecture to approximate all agents' joint state-action value. The mixing network non-linearly combines each agent's state-action value, which is only a condition on each agent's local observations. Furthermore, QMIX enforces the monotonic joint state-action value with each agent state-action value by employing a network structural monotonicity constraint to the mixing network.

However, these algorithms cannot handle continuous action space problems. The deep deterministic policy gradient (DDPG) \cite{silver2014deterministic} is a representative method in the continuous action task for a single agent. Multi-agent deep deterministic policy gradient (MADDPG) \cite{lowe2017multi} extended DDPG from single agent setting to the multi-agent setting by using centralized training and decentralized execution paradigm \cite{kraemer2016multi}. MADDPG is a representative algorithm for multi-agent continuous action problems. Thus, in this study, we model the TAPF in the intelligent warehouse as a cooperative multi-agent deep RL problem and use MADDPG to solve this problem simultaneously.

\section{Preliminaries}
\label{Preliminary}
% This section summarises the Markov decision process, RL, cooperative multi-agent deep RL, and the definition of MAPF and TAPF problems.

\subsection{Markov Decision Process and RL}
\label{Markov Decision Process and RL}
This study considers a finite-horizon Markov decision process \cite{sutton2018reinforcement}, defined by a tuple $(\mathcal{S}, \mathcal{A}, \mathcal{P}, r, \gamma, T)$. $\mathcal{S}$ denotes the state space, $\mathcal{A}$ represents the finite action space, $\mathcal{P}: \mathcal{S} \times \mathcal{A} \times \mathcal{S} \rightarrow [0,1]$ denotes the state transition distribution, $r: \mathcal{S} \times \mathcal{A} \rightarrow R$ denotes the reward function, $\gamma \in [0,1)$ denotes the discount factor and $T$ is a time horizon. At each time step $t$, an action $a_{t} \in A$ is chosen from a policy $\pi$. After transiting into the next state by sampling from $\mathcal{P}\left(s_{t+1} \mid s_{t}, a_{t}\right)$, the agent obtains a immediate reward $r\left(s_{t}, a_{t}\right)$. The agent continues performing actions until it enters a terminal state or $t$ reaches the time horizon $T$. RL aims to learn the policy $\pi: S \times A \rightarrow [0, 1] $ for decision-making problems by maximizing discounted cumulative rewards $\mathbb{E}_{\pi, \mathcal{P}}\left[r_{0: T-1}\right]=\mathbb{E}_{\pi, \mathcal{P}}\left[\sum_{0}^{T-1} \gamma^{t} r\left(s_{t}, a_{t}\right)\right]$. In this study, we cannot access the environment dynamics $\mathcal{P}$, which means model-free deep RL learning.

\subsection{Cooperative Multi-Agent Deep RL}
\label{Multi-Agent Deep RL}
A fully cooperative multi-agent problem can be modeled as a Markov decision process \cite{sutton2018reinforcement}, described by a tuple $G=(\mathcal{S}, \mathcal{A}, \mathcal{P}, r, \mathcal{O}, N, \gamma, T)$. $s \in S$ represents the true state. At each timestep $t$, each agent $n \in N:=\{1, \ldots, N\}$ selects an action $a^n_{t} \in \mathcal{A}$, then the joint action $\boldsymbol{a_t} = \{a^{1}_t, a^{2}_t, ..., a^{N}_t\}$ is obtained. $\mathcal{P}\left(s^{\prime}_{t+1} \mid s_t, \boldsymbol{a_t} \right): \mathcal{S} \times \mathcal{A}^{N} \times S \rightarrow [0,1]$ denotes the state transition function. $r(s_t, \boldsymbol{a_t}): \mathcal{S} \times \mathcal{A}^{N} \rightarrow R$ denotes the reward function shared by all agents, $\gamma \in[0,1)$ represents the discount factor, and $T$ denotes the time horizon. Each agent $n$ has an independent observation $o \in \mathcal{O}$ and a history of action observation $\tau^{n} \in \mathcal{T} \equiv(\mathcal{O} \times \mathcal{A})$. A stochastic policy $\pi^{n}\left(a^{n} \mid \tau^{n}\right): \mathcal{T} \times A \rightarrow[0,1]$ is based on the history of action-observation. Let $\pi^{n}$ denote agent $n$' policy; cooperative agents aim to maximize:
\begin{equation}
    J(\pi)=\mathbb{E}_{s_{0:T-1}, a^{1} \sim \pi^{1}, \ldots, a^{N} \sim \pi^{N}}\left[\sum_{t=0}^{T-1} \gamma^{t} r\left(s_{t}, a_{t}^{1}, \ldots, a_{t}^{N}\right)\right]
    \label{eq 1}
\end{equation}

In cooperative multi-agent deep RL, the training process is centralized, but the execution process is decentralized \cite{kraemer2016multi}. This means that, in the learning process, each agent has access to the global state $s$. However, each agent's decision only depends on its own action-observation history $\tau^{n}$ in the execution process.

\begin{figure}[htbp]
    \centering
    \subfloat[Five agents - five tasks]{  
    \begin{minipage}[b]{0.24\textwidth} 
        \centering
        \includegraphics[width=4cm,height=4cm]{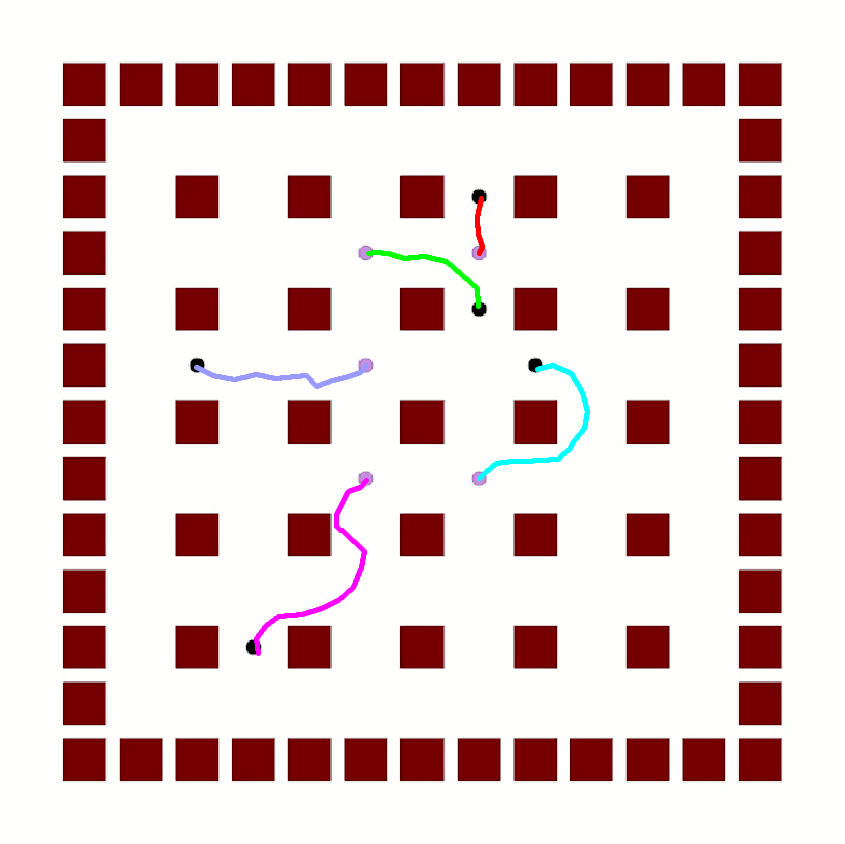} 
    \end{minipage}}
    \subfloat[Five agents - twenty tasks]{ 
    \begin{minipage}[b]{0.24\textwidth} 
        \centering
        \includegraphics[width=4cm,height=4cm]{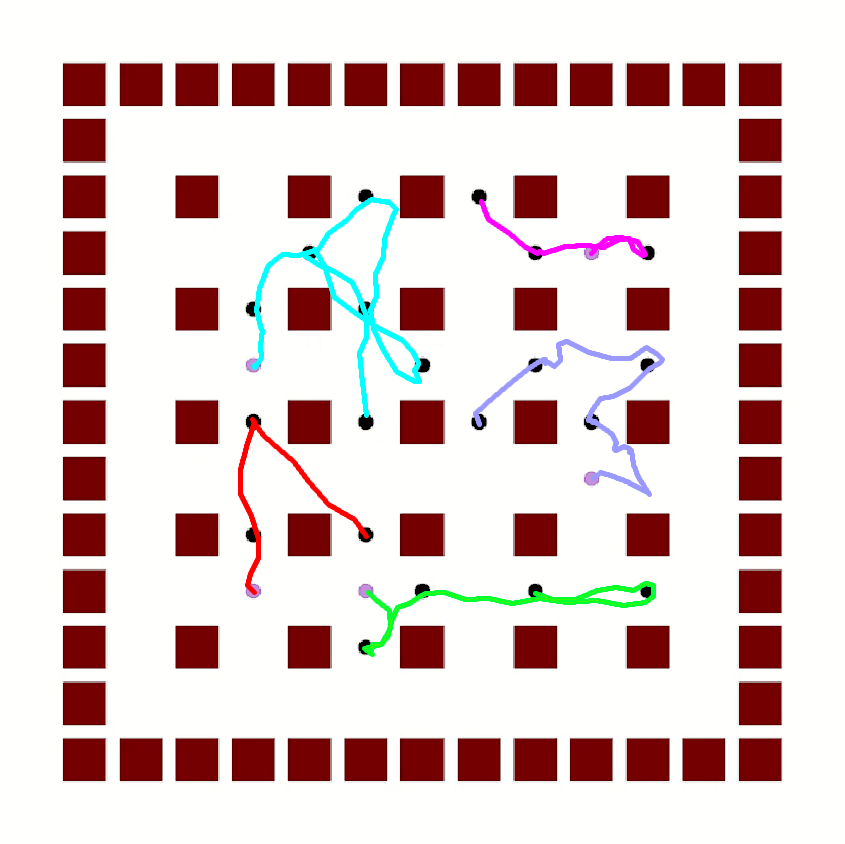} 
    \end{minipage}}
    \caption{Modeling TAPF as a MARL problem}
    \label{Figures 1} 
\end{figure}

\begin{figure*}[hbp]
	\centering
	\includegraphics[width=11cm,height=4cm]{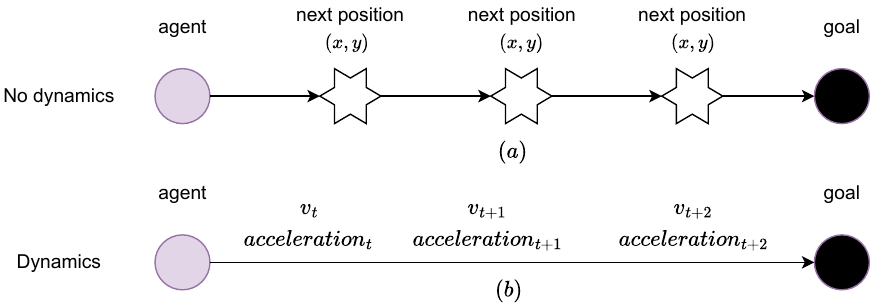}
	\caption{The physical dynamics of agents} 
	\label{Figures 2}
\end{figure*}

\subsection{Definitions of MAPF and TAPF Problems}
\label{Defination of MAPF and TAPF problem}
MAPF problem: In the multi-agent path finding (MAPF) problem, the input includes an environment consisting of N agents. The environment defines all possible start and goal positions for the agents. Each agent \( n \) has its own start position \( start^n \) and goal position \( goal^n \). At time \( t \), agent \( n \) is at a specific location and can execute an action \( a^n_t \), subsequently moving to a new position or staying at the current position. The action can be represented as \( a^n_t: \text{current position} \rightarrow \text{new position} \). The path \( \pi_n \) of agent \( n \) consists of a sequence of actions \( \pi_n = (a^n_1, a^n_2, \ldots, a^n_{T-1}) \). Conflicts in MAPF typically involve position and movement collisions between agents.

TAPF problem: The TAPF problem also involves $N$ agents within a given environment. Unlike MAPF, each agent's goal position is not predefined. The task set includes $m$ target positions. Each agent \( n \) has a defined start position, but its goal positions are dynamically assigned based on the task set. The objective of TAPF is to define a goal position assignment scheme for each intelligent system to minimize the sum of the paths of all agents. Time is also considered discrete in this problem, with agents deciding their actions at each timestep based on their current position and assigned goals.

\section{Target Assignment and Path Finding for Intelligent Warehouse: A Cooperative Multi-Agent Deep RL Perspective}
\label{Multi-Agent Target Assignment and Path Finding for the Intelligent Warehouse: A Multi-agent Deep Reinforcement Learning Perspective}
In Section \ref{Modeling TAPF as a MARL Problem}, we detailedly describe our method that models the TAPF problem as a cooperative multi-agent RL problem. In Section \ref{Solving TAPF via MARL}, we introduce the MADDPG algorithm to solve TAPF.

\subsection{Modeling TAPF as a MARL Problem}
\label{Modeling TAPF as a MARL Problem}

As shown in Fig. \ref{Figures 1}, we provide two scenarios to describe how we model the TAPF problem (described in Section \ref{Defination of MAPF and TAPF problem}) as a cooperative MARL problem. As introduced in Section \ref{Multi-Agent Deep RL}, cooperative multi-agent deep RL can be modeled as a tuple $G=(\mathcal{S}, \mathcal{A}, \mathcal{P}, r, \mathcal{O}, N, \gamma, T)$. In this study, the state transition function $\mathcal{P}\left(s^{\prime}_{t+1} \mid s_t, \boldsymbol{a_t} \right): \mathcal{S} \times \mathcal{A}^{N} \times \mathcal{S} \rightarrow [0,1]$ is unknown, which means the agent cannot get the environment dynamics. This is consistent with the real-world TAPF problem. The number of agents $N$ in this study can be set to any integer number. The discount factor $\gamma=0.99$ is the most common setting in deep RL. Since the main elements in multi-agent deep RL are the observation space $\mathcal{O}$ (or state space $\mathcal{S}$), the action space $\mathcal{A}$, and the reward function $r$. Thus, we describe these elements in detail.

\textbf{Observation space:} An agent's observation contains the position and velocity itself, the relative positions of all tasks, the relative positions of the other agents, and the relative positions of neighboring obstacles. In this study, the number of other visible agents in an agent's observation can be set to equal to or less than $N-1$. Taking Fig. \ref{Figures 1} (a) as an example, the five purple circles represent five agents, and the five black circles represent five tasks. For obstacles, we set the agent to perceive the relative positions of the adjacent obstacles, and the number of adjacent obstacles can be set to any integer. In Fig. \ref{Figures 1} (a) and (b), different color curves denote the navigation trajectories of different agents.

\begin{figure}[htbp]
	\centering
	\includegraphics[width=7cm,height=5.5cm]{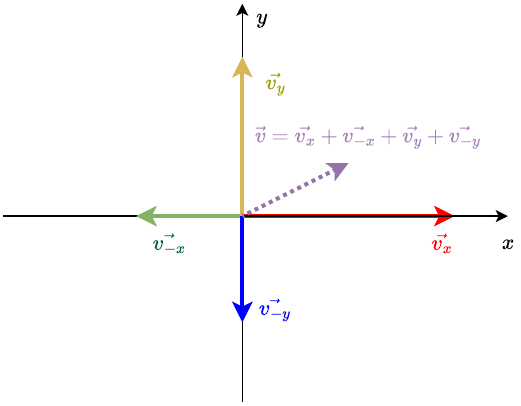}
	\caption{Action space of an agent} 
	\label{Figures 3}
\end{figure}

\textbf{The physical dynamics of agents:} Before describing the action space, we first describe the dynamics of agents. Note that our work is the first to consider the dynamics of agents in TAPF for the intelligent warehouse. As shown in Fig. \ref{Figures 2} (a), the "No dynamics" schematic diagram represents previous literature \cite{sartoretti2019primal,9366340} methods that only focus on the next position of an agent but do not consider velocities and accelerations of agents. Further, the dynamics of agents in previous literature are solved by using traditional methods \cite{ROSplanning}. On the contrary, the "Dynamics" schematic diagram in Fig. \ref{Figures 2} (b) represents our method that considers the dynamics of agents. We can calculate agents' velocities and accelerations.

The output of our policy network is the magnitude and direction of four cardinal directions force ($\vec{F}_x, \vec{F}_{-x}, \vec{F}_y, \vec{F}_{-y}$) applied to the agent. According to Newton's second law of motion, $\vec{F}_{direction}=m*\vec{a}_{direction}$ (where $m$ is the mass of the agent), we can get the acceleration $\vec{a}_{direction}=\frac{\vec{F}_{direction}}{m}$. According to basic physics knowledge, we can get the velocity of the agent $\vec{v}_{t(direction)} = \vec{v}_{0(direction)} + \vec{a}_{direction}* \Delta t$, where $v_0$ is the initial velocity, and $\Delta t$ denotes a time interval.

\textbf{Action space:} In this study, the action space of an agent is continuous, representing the movements. As shown in Fig. \ref{Figures 3}, an agent obtains a velocity between 0.0m/s and 1.0m/s in each of the four cardinal directions \textit{[move\_left ($\vec{v}_{-x}$), move\_right ($\vec{v}_x$), move\_down ($\vec{v}_{-y}$), move\_up ($\vec{v}_{y}$)]}, the final action ($\vec{v}$) is calculated by the vector sum of the four velocities.

\textbf{Reward function:} 
The reward function is defined as:
\begin{equation}
	\begin{aligned}
		& r = reward\_success + reward\_distance\_tasks\_to\_agents \\
		& \quad + reward\_collision\_agents\_to\_obstacles \\
		& \quad  + reward\_collision\_agents\_to\_agents
		\label{eq 2}
	\end{aligned}	
\end{equation}

\begin{figure}[htbp]
	\centering
	\includegraphics[width=8cm,height=3.5cm]{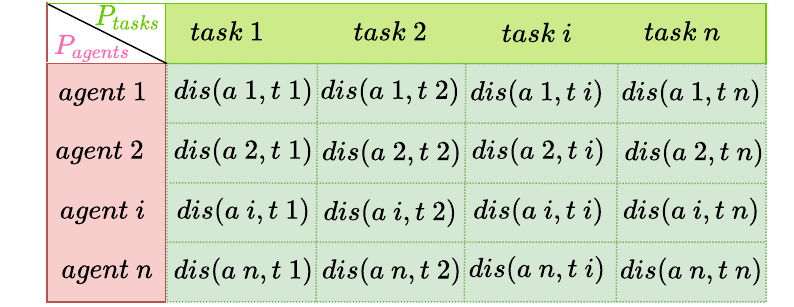}
	\caption{Distansce matrix of tasks and agents: $\Vert \boldsymbol{P_{tasks} - P_{agents}} \Vert$. The element $dis(a \ i, t  \ i)$ represents the distance between $agent \ i$ and $task  \ i$.} 
	\label{Figures 4}
\end{figure}

\noindent The detailed rewards are defined as follows ($R$ is the radius of an agent):
\begin{itemize}
	\item reward\_success = $100*n$, \\
	$\text{where}\,n\,\text{denotes the number of elements less than 0.05 in } \\
	 \Vert \boldsymbol{P_{tasks} - P_{agents}} \Vert$.
	\item reward\_distance\_tasks\_to\_agents \\
	= -$\Vert \boldsymbol{P_{tasks} - P_{agents}} \Vert_{min}$
	\item reward\_collision\_agents\_to\_obstacles = -$2*n$, \\
	$\text{where}\,n\,\text{denotes the number of elements less than $R$ in } \\
	\Vert \boldsymbol{P_{obstacles} - P_{agents}} \Vert$.
	\item reward\_collision\_agents\_to\_agents = -$2*n$, \\
	$\text{where}\,n\,\text{denotes the number of elements less than $2R$ in } \\
	\Vert \boldsymbol{P_{agents} - P_{agents}} \Vert$.
\end{itemize}

For reward\_success, $\Vert \boldsymbol{P_{tasks} - P_{agents}} \Vert$ represents the distance matrix of all tasks to all agents. Fig. \ref{Figures 4} shows a detailed distance matrix. If the distance of $agent \ i$ to $task \ i$ is less than $0.05m$, we give a $+100$ positive reward to force agents to navigate to the tasks. If there are $n$ elements in $\Vert \boldsymbol{P_{tasks} - P_{agents}} \Vert$ less than $0.05m$, we set reward\_success = +$100*n$. Reward\_distance\_tasks\_to\_agents is created to provide a dense reward to accelerate the training speed. In $\Vert \boldsymbol{P_{tasks} - P_{agents}} \Vert_{min}$, the subscript $min$ means that we sum the minimum distance of each task to every agent. 

Reward\_collision\_agents\_to\_obstacles aims to punish agents from collisions with obstacles. The definition of $\Vert \boldsymbol{P_{obstacles} - P_{agents}} \Vert$ is similar with $\Vert \boldsymbol{P_{tasks} - P_{agents}} \Vert$, $\Vert \boldsymbol{P_{obstacles} - P_{agents}} \Vert$ denotes the distance matrix of obstacles and tasks. If the distance of an agent to an obstacle is less than $R$, a negative reward $-2$ is given. If there are $n$ elements in $\Vert \boldsymbol{P_{obstacles} - P_{agents}} \Vert$ less than $R$, we set reward\_collision\_agents\_to\_obstacles = -$2*n$. 

Reward\_collision\_agents\_to\_agents aims to punish agents from collision to agents. The definition of $\Vert \boldsymbol{P_{agents} - P_{agents}} \Vert$ is similar with $\Vert \boldsymbol{P_{tasks} - P_{agents}} \Vert$, $\Vert \boldsymbol{P_{agents} - P_{agents}} \Vert$ denotes the distance matrix of agents and agents. If the distance of an agent to an agent is less than $2R$, we give a negative reward $-2$. If there are $n$ elements in $\Vert \boldsymbol{P_{agents} - P_{agents}} \Vert$ less than $2R$, we set reward\_collision\_agents\_to\_agents = -$2*n$.

\subsection{Solving TAPF Problem via Cooperative MARL}
\label{Solving TAPF via MARL}
In this study, we introduce MADDPG \cite{lowe2017multi} to solve the TAPF problem. Considering that agents are homogeneous, they can share the same policy network. This makes learning policy more efficient. Let $N$ denote the number of agents, and $\pi$ denote the policy parameterized by $\theta$. The centralized critic $Q$ is parameterized by $\phi$. The parameters $\theta$ for policy $\pi$ can be updated iteratively by the associated critic $Q$:
\begin{equation}
\begin{aligned}
\nabla_{\theta} J\left(\theta\right) & =\mathbb{E}_{\left(\boldsymbol{x_{t}}, \boldsymbol{a_{t}}\right) \sim \mathcal{D}}\Big[\nabla_{\theta} \pi\left(o_{t}^{n} \mid \theta \right) \\
& \quad \nabla_{a_{t}} Q\left(\boldsymbol{x_{t}}, a_{t}^{1}, \ldots, a_{t}^{N} \mid \phi \right)\vert_{a_{t}^{n}=\pi\left(o_{t}^{n} \mid \theta \right)}\Big]
\end{aligned}
\label{eq 3}
\end{equation}
where $\boldsymbol{x}_{t}$ and $\boldsymbol{a}_{t}$ are the concatenation of all agents' observation and action at time step $t$, $\boldsymbol{x}_{t}=\left(o_{t}^{1}, \ldots, o_{t}^{N}\right)$ and $\boldsymbol{a}_{t}=\left(a_{t}^{1}, \ldots, a_{t}^{N}\right)$. $o_{t}^{n}$ is the observation received by agent $n$ at time step $t$. $\mathcal{D}$ denotes the replay buffer containing tuples ($\boldsymbol{x}_{t}, \boldsymbol{a}_{t}, r_{t} \cdots$).

The centralized critic parameters $\phi$ are optimized by minimizing the loss:
\begin{equation}
	L\left(\phi\right)=\mathbb{E}_{\left(\boldsymbol{x}_{t}, \boldsymbol{a}_{t}, {r}_{t}, \boldsymbol{x}_{t+1}\right) \sim \mathcal{D}}\left[\left(Q\left(\boldsymbol{x}_{t}, \boldsymbol{a}_{t} \mid \phi\right)-y_{t}\right)^{2}\right]
	\label{eq 4}
\end{equation}
where ${r}_{t}$ is the concatenation of rewards (Eq (\ref{eq 2})) received by all agents at time step $t$. The target critic value $y_{t}^{n}$ is defined as:
\begin{equation}
	y_{t}=r_{t}+\gamma Q^{-}\left(\boldsymbol{x}_{t+1}, a_{t+1}^{1}, \ldots, a_{t+1}^{N} \mid \phi^{-} \right)\vert_{a_{t+1}^{n}=\pi\left(o_{t+1}\right), n=1, \ldots, N}
	\label{eq 5}
\end{equation}
where $Q^{-}$ is a target Q-network parameterized by $\phi^{-}$. $\phi^{-}$ is optimized by:
\begin{equation}
	\phi^{-} = \alpha \phi + (1-\alpha) \phi^{-}
	\label{eq 6}
\end{equation}
where $\alpha$ is a coefficient to trade off the weight of Q-network and target Q-network. 

The shared reward makes the agents learn policy cooperatively. In the execution phase, we only use the policy $\pi$ whose inputs are $o_{t}^{1}, \ldots, o_{t}^{N}$, respectively, and the outputs are force ($\boldsymbol{\vec{F}}$) applied to the agent. According to the physical dynamics of agents described in Section \ref{Modeling TAPF as a MARL Problem}, we can get the actions $\boldsymbol{a}_{t}=\left(a_{t}^{1}, \ldots, a_{t}^{N}\right)$. Note that, for agent $n$, the input of $\pi$ is only the observation $o_{t}^{n}$ received by agent $n$ at time step $t$ in the execution phase.

\begin{figure}[htbp]
    \centering
    \subfloat[Two agents - two tasks]{  
    \begin{minipage}[b]{0.24\textwidth} 
        \centering
        \includegraphics[width=4cm,height=3.8cm]{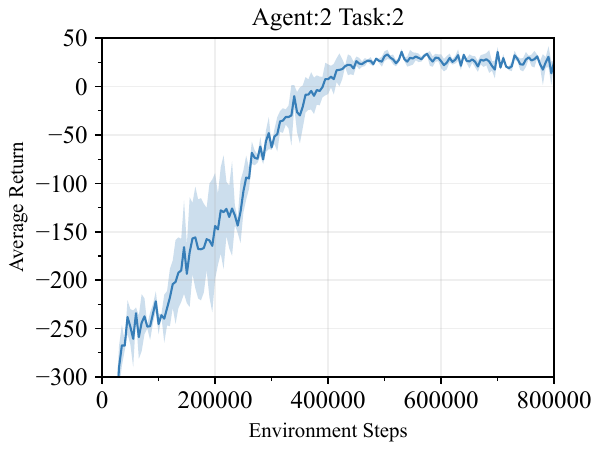} 
    \end{minipage}}
    \subfloat[Two agents - four tasks]{ 
    \begin{minipage}[b]{0.24\textwidth} 
        \centering
        \includegraphics[width=4cm,height=3.8cm]{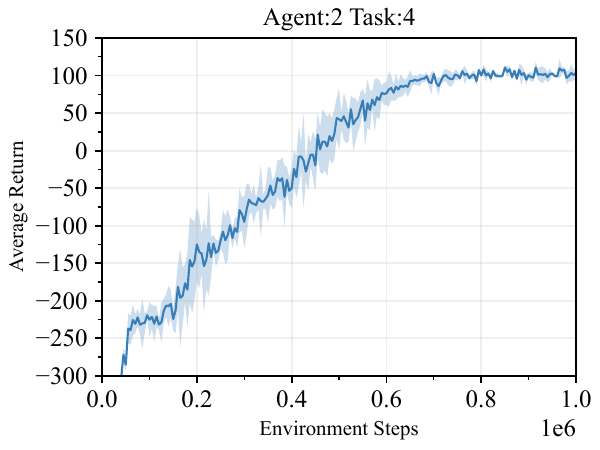} 
    \end{minipage}}

    \subfloat[Five agents - five tasks]{  
    \begin{minipage}[b]{0.24\textwidth} 
        \centering
        \includegraphics[width=4cm,height=3.8cm]{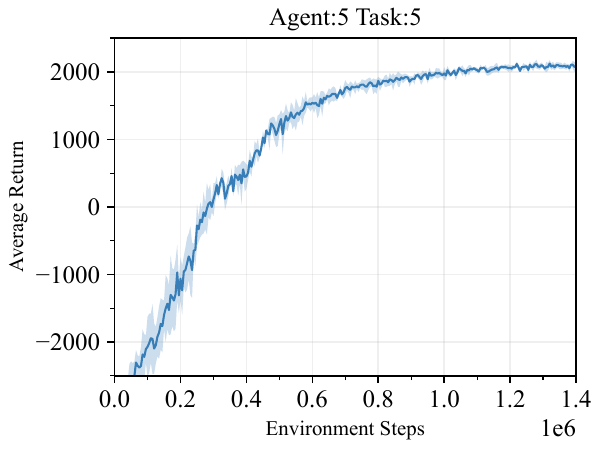} 
    \end{minipage}}
    \subfloat[Five agents - ten tasks]{ 
    \begin{minipage}[b]{0.24\textwidth} 
        \centering
        \includegraphics[width=4cm,height=3.8cm]{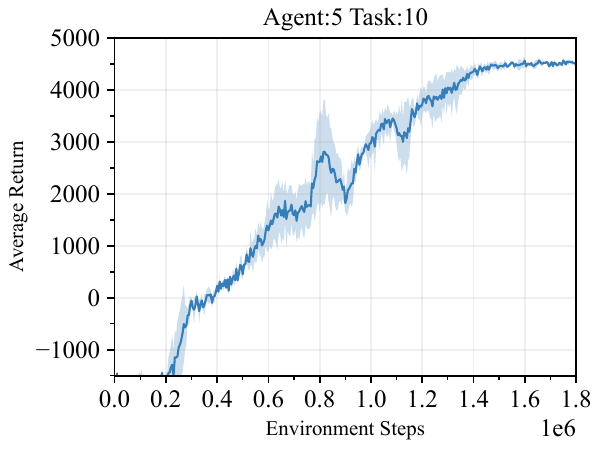} 
    \end{minipage}}

    \subfloat[Five agents - twenty tasks]{  
    \begin{minipage}[b]{0.24\textwidth} 
        \centering
        \includegraphics[width=4cm,height=3.8cm]{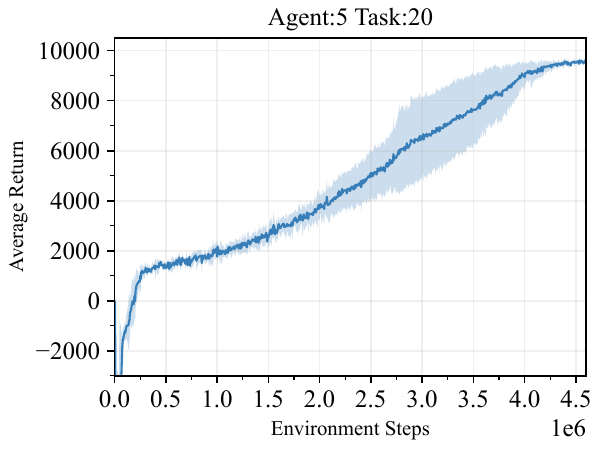} 
    \end{minipage}}
    \caption{Average return}
    \label{Figures 5} 
\end{figure}

\section{Experiments}
\label{Experiments}
In this section, we verified the proposed method from three aspects. In Section \ref{Target assignment and path planning}, we verified the proposed method's target assignment and path planning performances. In all the experiments, the locations of agents and task points are generated randomly. Section \ref{Cooperation ability} verified the learned cooperation ability. Section \ref{Time efficiency} verified the time efficiency. 

\begin{figure}[htbp]
    \centering
    \subfloat[]{  
    \begin{minipage}[b]{0.24\textwidth} 
        \centering
        \includegraphics[width=4cm,height=4cm]{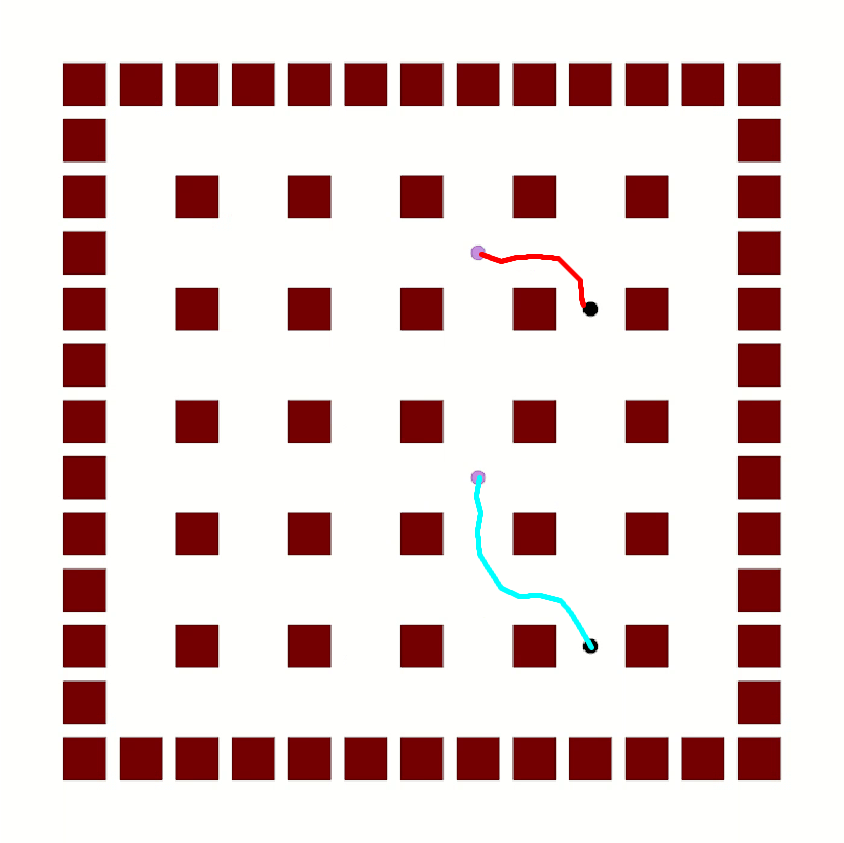} 
    \end{minipage}}
    \subfloat[]{ 
    \begin{minipage}[b]{0.24\textwidth} 
        \centering
        \includegraphics[width=4cm,height=4cm]{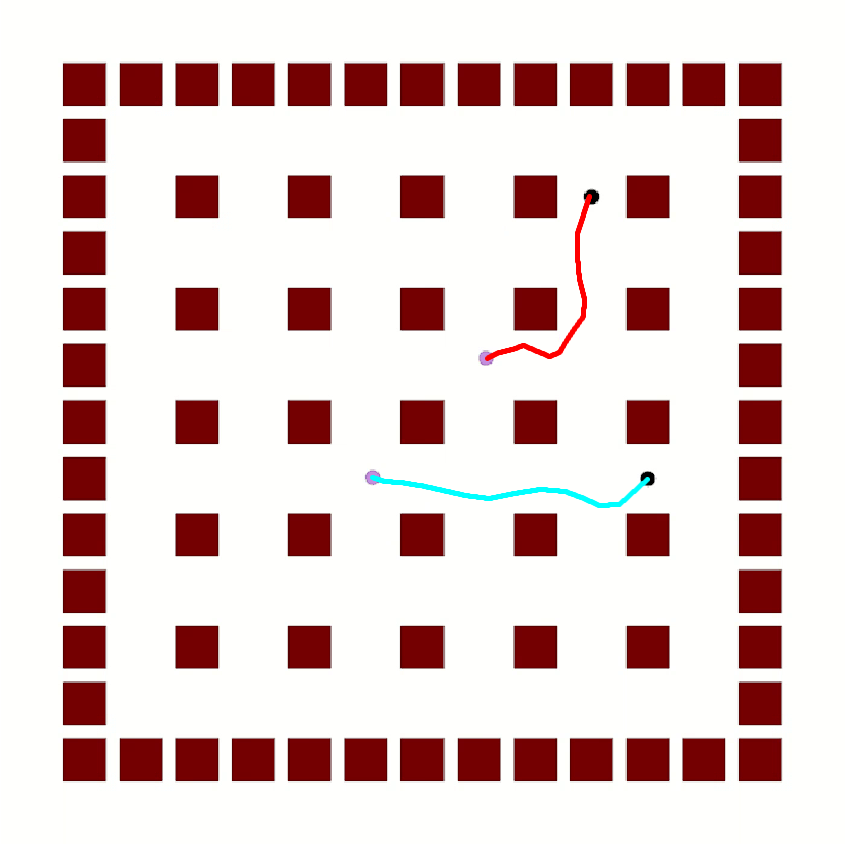} 
    \end{minipage}}

    \subfloat[]{  
    \begin{minipage}[b]{0.24\textwidth} 
        \centering
        \includegraphics[width=4cm,height=4cm]{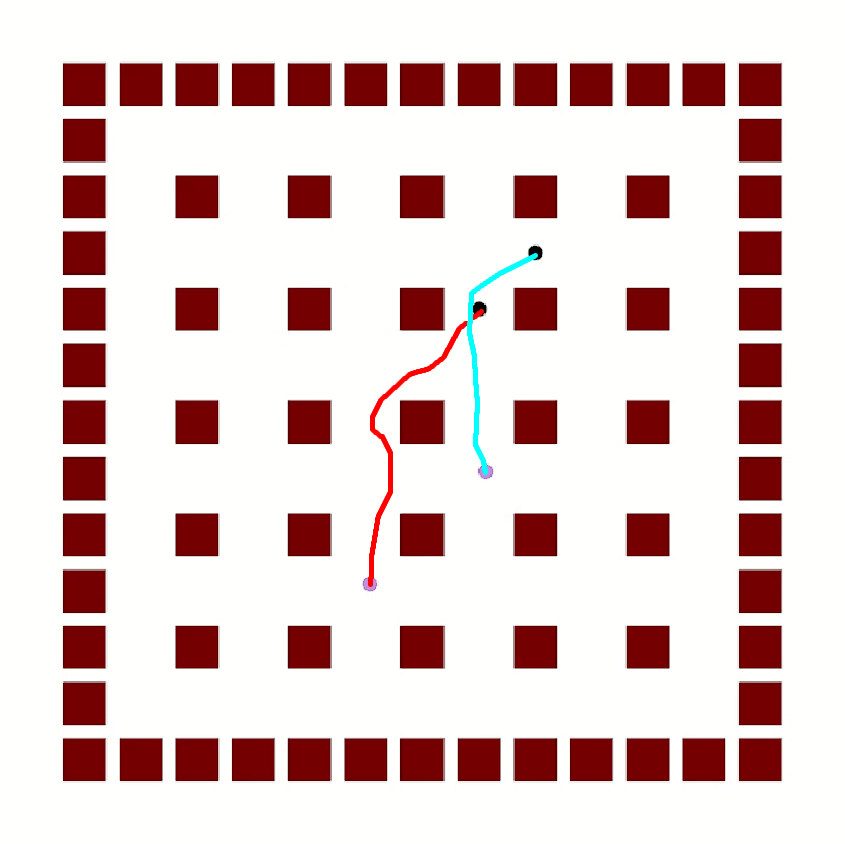} 
    \end{minipage}}
    \subfloat[]{ 
    \begin{minipage}[b]{0.24\textwidth} 
        \centering
        \includegraphics[width=4cm,height=4cm]{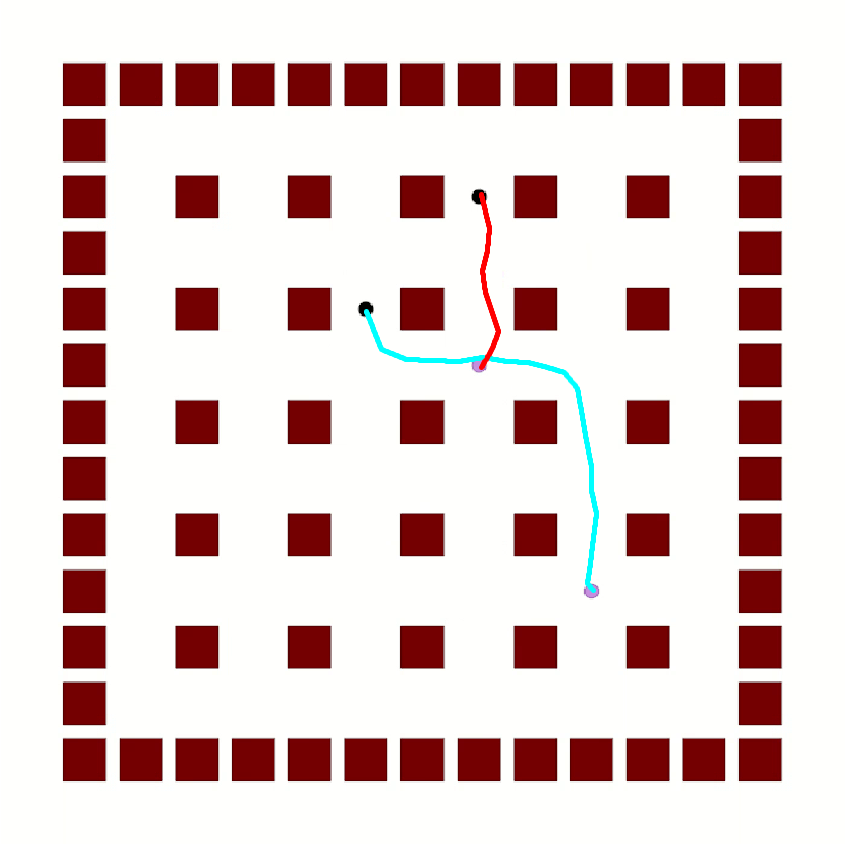} 
    \end{minipage}}
    \caption{Two agents - two tasks}
    \label{Figures 6} 
\end{figure}

\subsection{Target Assignment and Path Planning}
\label{Target assignment and path planning}
This subsection verified the proposed method in various intelligent warehouse settings. We set five different level scenarios to verify the performances: (1) \textit{two agents - two tasks} (2) \textit{two agents - four tasks} (3) \textit{five agents - five tasks} (4) \textit{five agents - ten tasks} (5) \textit{five agents - twenty tasks}.

\begin{figure}[htbp]
    \centering
    \subfloat[]{  
    \begin{minipage}[b]{0.24\textwidth} 
        \centering
        \includegraphics[width=4cm,height=4cm]{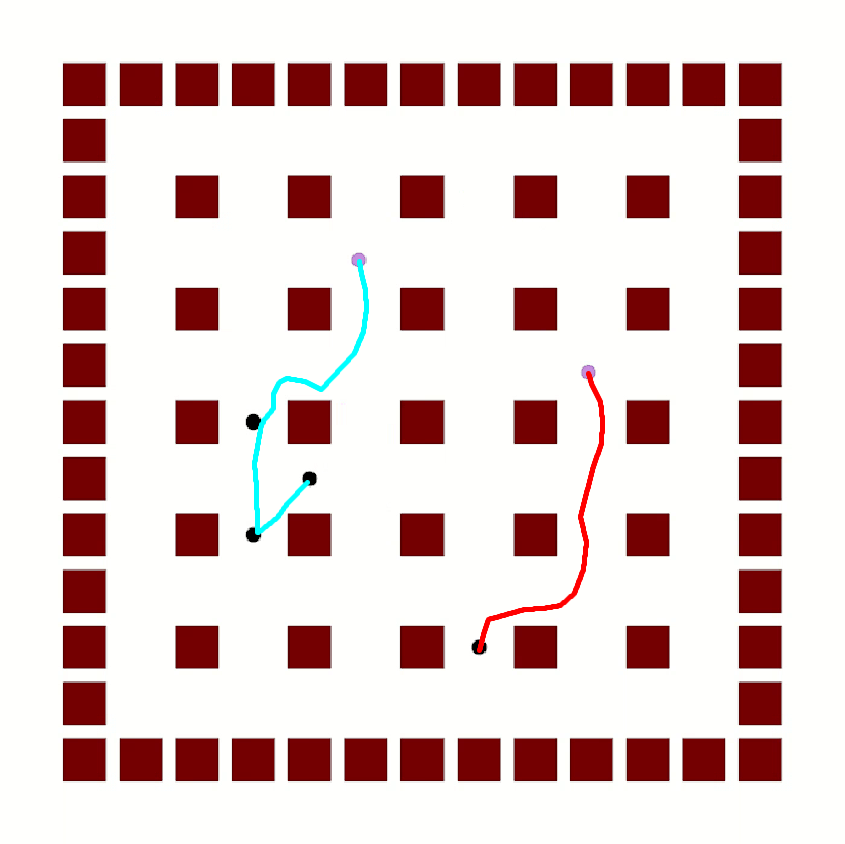} 
    \end{minipage}}
    \subfloat[]{ 
    \begin{minipage}[b]{0.24\textwidth} 
        \centering
        \includegraphics[width=4cm,height=4cm]{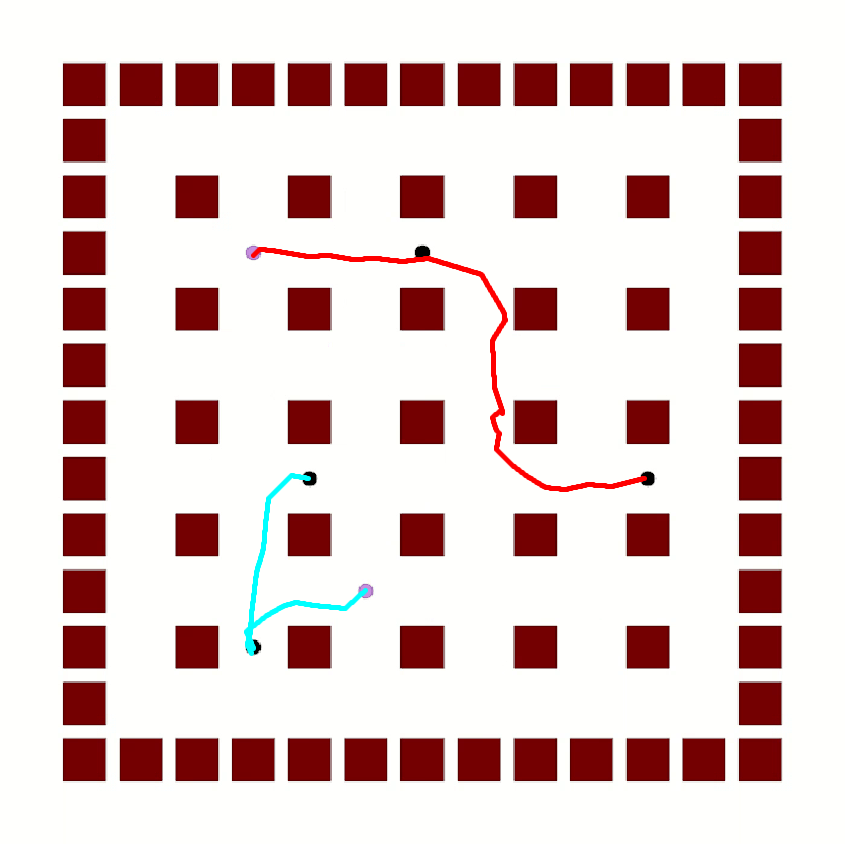} 
    \end{minipage}}

    \subfloat[]{  
    \begin{minipage}[b]{0.24\textwidth} 
        \centering
        \includegraphics[width=4cm,height=4cm]{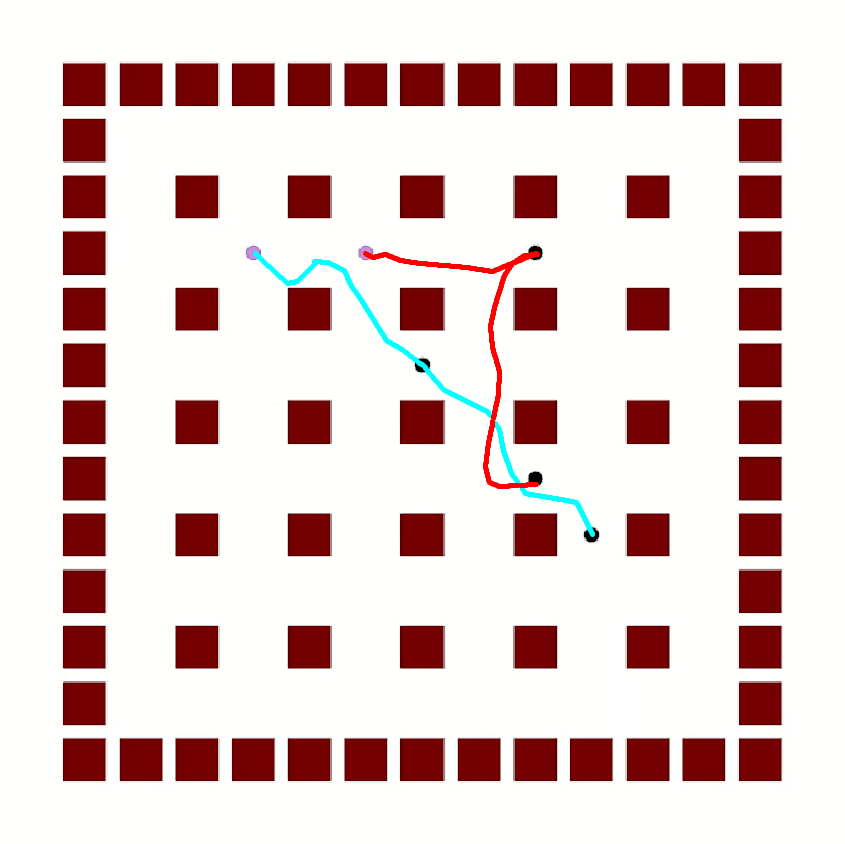} 
    \end{minipage}}
    \subfloat[]{ 
    \begin{minipage}[b]{0.24\textwidth} 
        \centering
        \includegraphics[width=4cm,height=4cm]{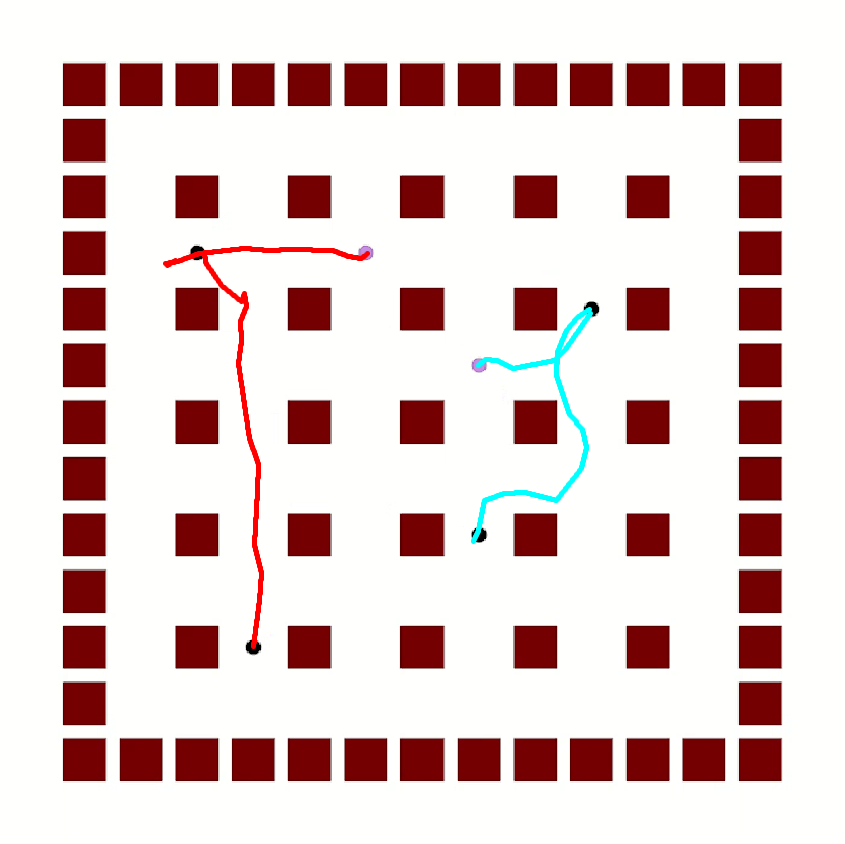} 
    \end{minipage}}
    \caption{Two agents - four tasks}
    \label{Figures 7} 
\end{figure}

\begin{figure}[htbp]
    \centering
    \subfloat[]{  
    \begin{minipage}[b]{0.24\textwidth} 
        \centering
        \includegraphics[width=4cm,height=4cm]{figures/a5_l5/a5_l5_r3_4.png} 
    \end{minipage}}
    \subfloat[]{ 
    \begin{minipage}[b]{0.24\textwidth} 
        \centering
        \includegraphics[width=4cm,height=4cm]{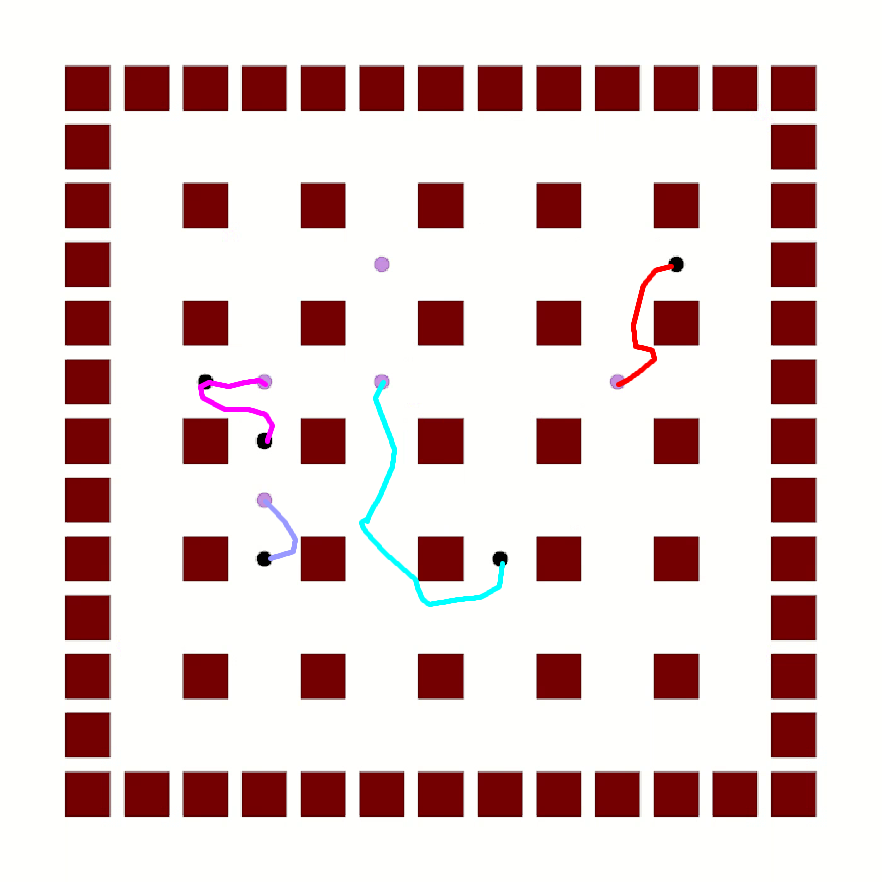} 
    \end{minipage}}

    \subfloat[]{  
    \begin{minipage}[b]{0.24\textwidth} 
        \centering
        \includegraphics[width=4cm,height=4cm]{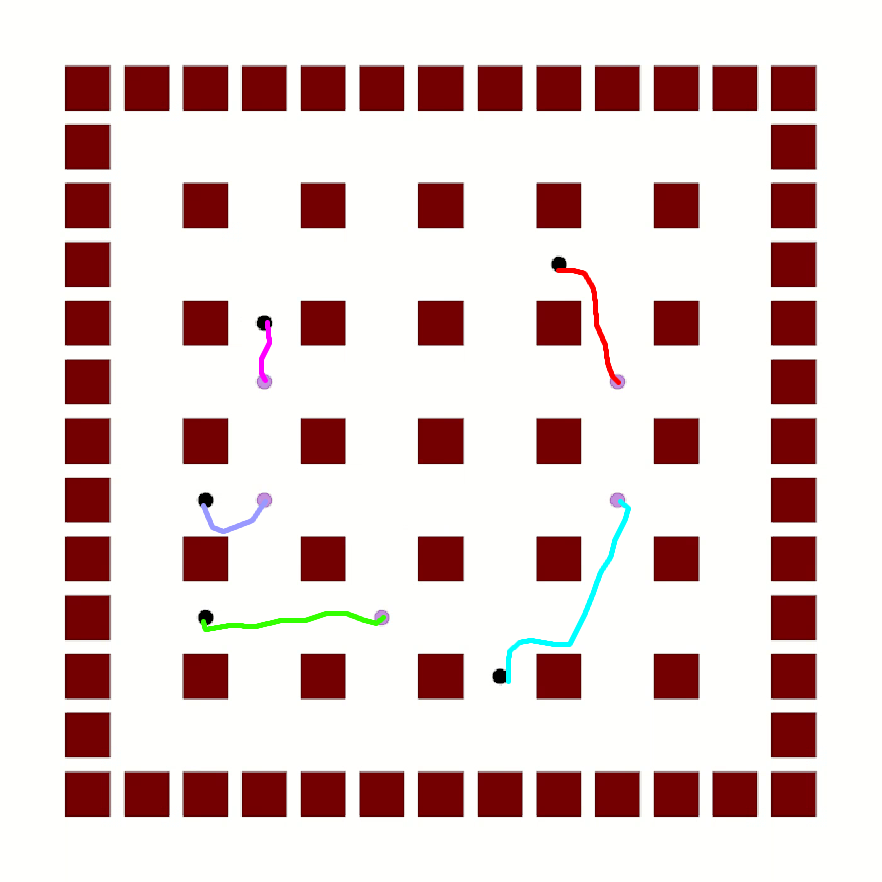} 
    \end{minipage}}
    \subfloat[]{ 
    \begin{minipage}[b]{0.24\textwidth} 
        \centering
        \includegraphics[width=4cm,height=4cm]{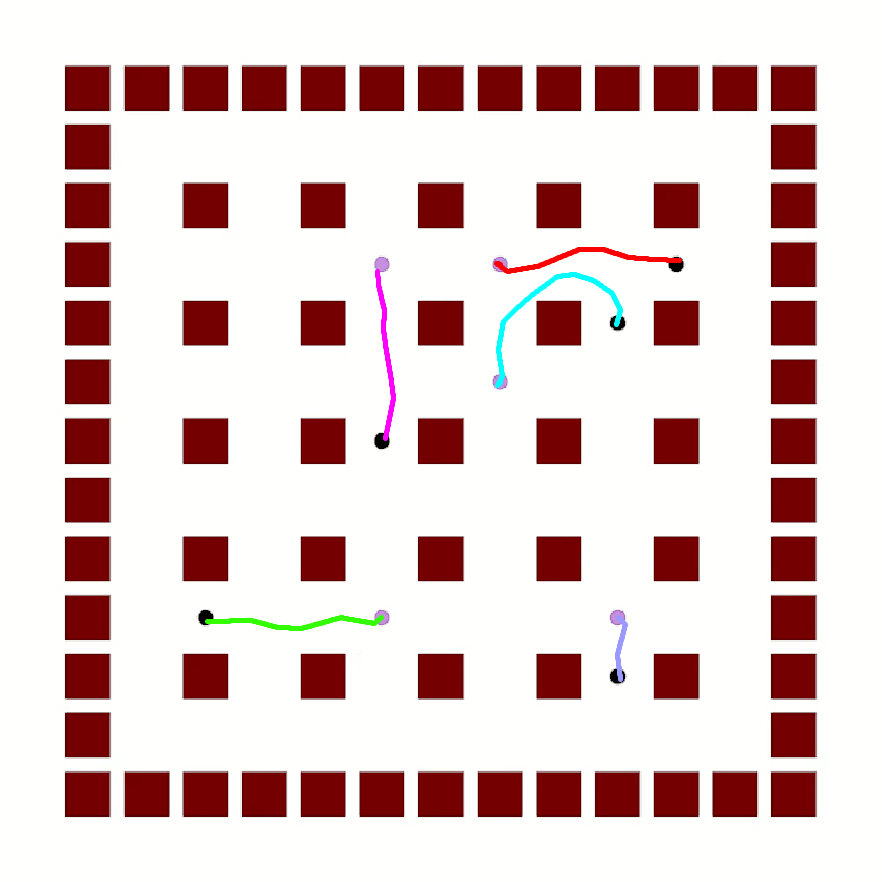} 
    \end{minipage}}
    \caption{Five agents - five tasks}
    \label{Figures 8} 
\end{figure}

\begin{figure}[htbp]
    \centering
    \subfloat[]{  
    \begin{minipage}[b]{0.24\textwidth} 
        \centering
        \includegraphics[width=4cm,height=4cm]{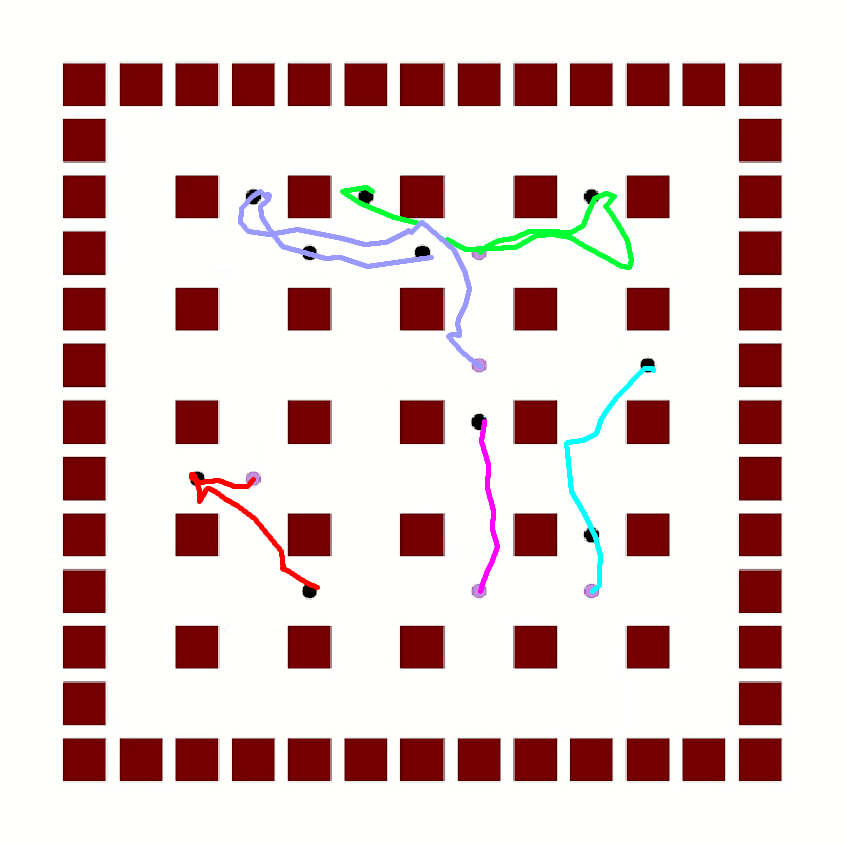} 
    \end{minipage}}
    \subfloat[]{ 
    \begin{minipage}[b]{0.24\textwidth} 
        \centering
        \includegraphics[width=4cm,height=4cm]{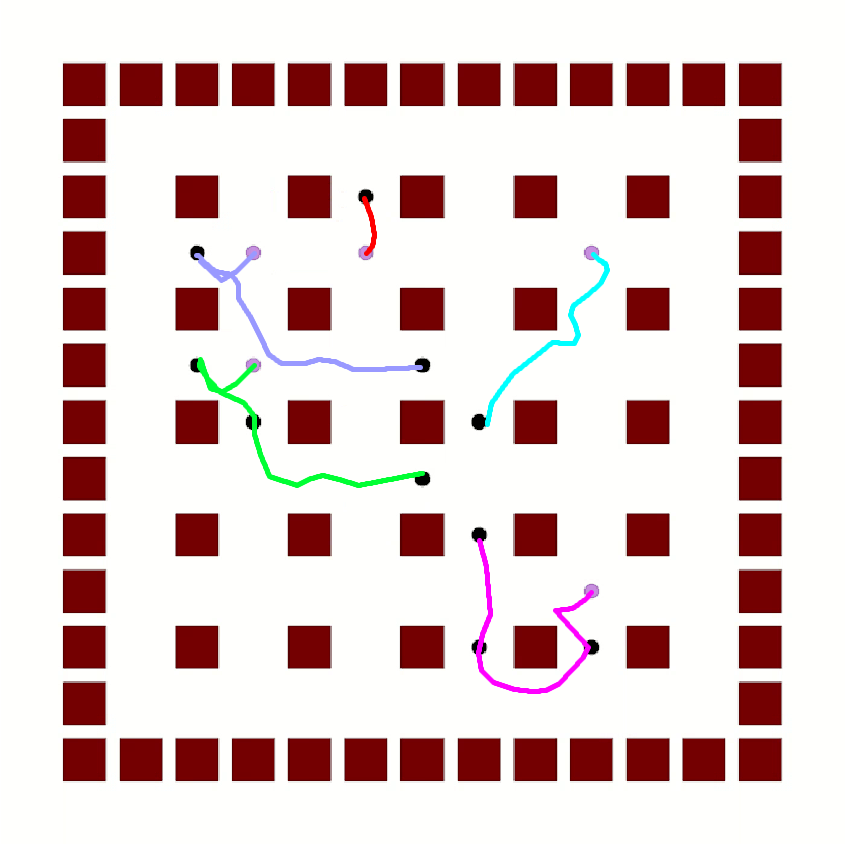} 
    \end{minipage}}

    \subfloat[]{  
    \begin{minipage}[b]{0.24\textwidth} 
        \centering
        \includegraphics[width=4cm,height=4cm]{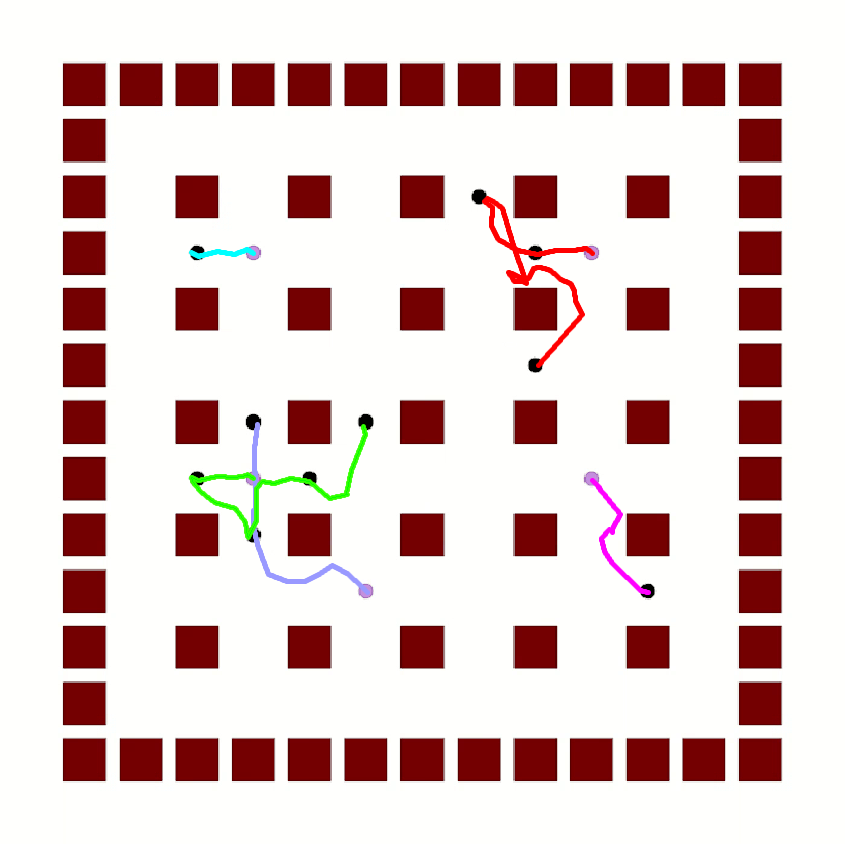} 
    \end{minipage}}
    \subfloat[]{ 
    \begin{minipage}[b]{0.24\textwidth} 
        \centering
        \includegraphics[width=4cm,height=4cm]{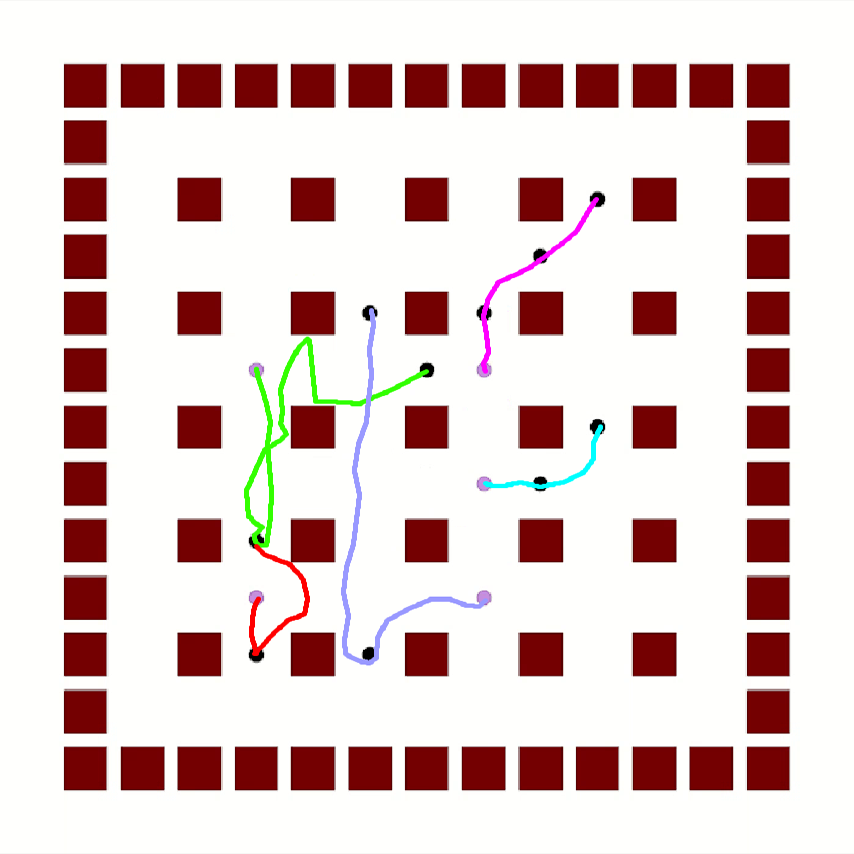} 
    \end{minipage}}
    \caption{Five agents - ten tasks}
    \label{Figures 9} 
\end{figure}

\begin{figure}[htbp]
    \centering
    \subfloat[]{  
    \begin{minipage}[b]{0.24\textwidth} 
        \centering
        \includegraphics[width=4cm,height=4cm]{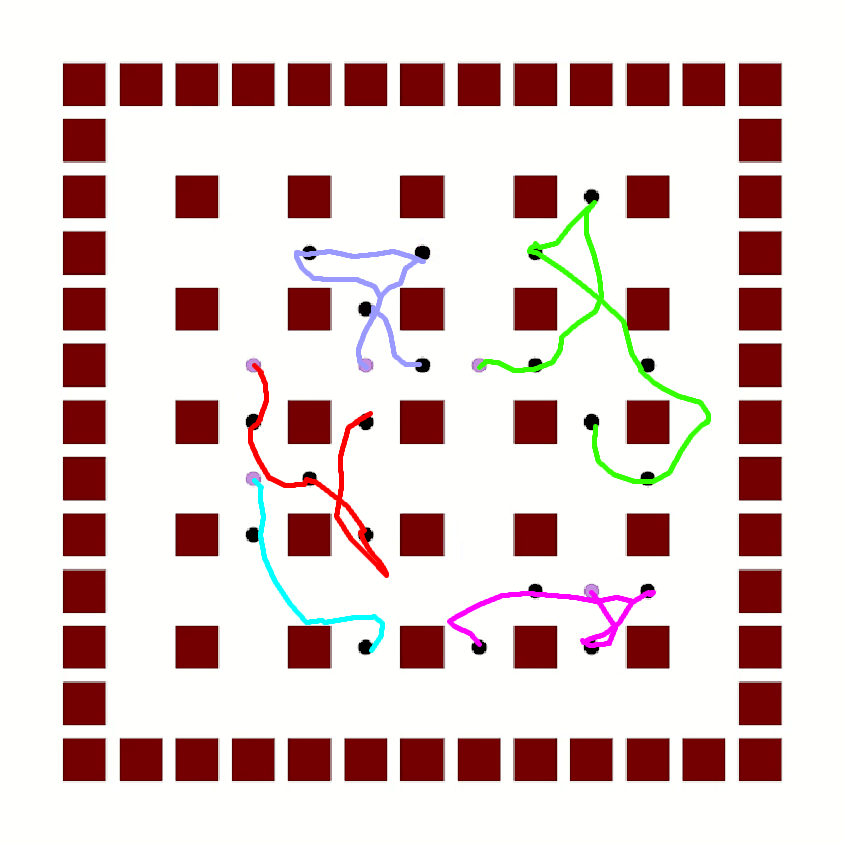} 
    \end{minipage}}
    \subfloat[]{ 
    \begin{minipage}[b]{0.24\textwidth} 
        \centering
        \includegraphics[width=4cm,height=4cm]{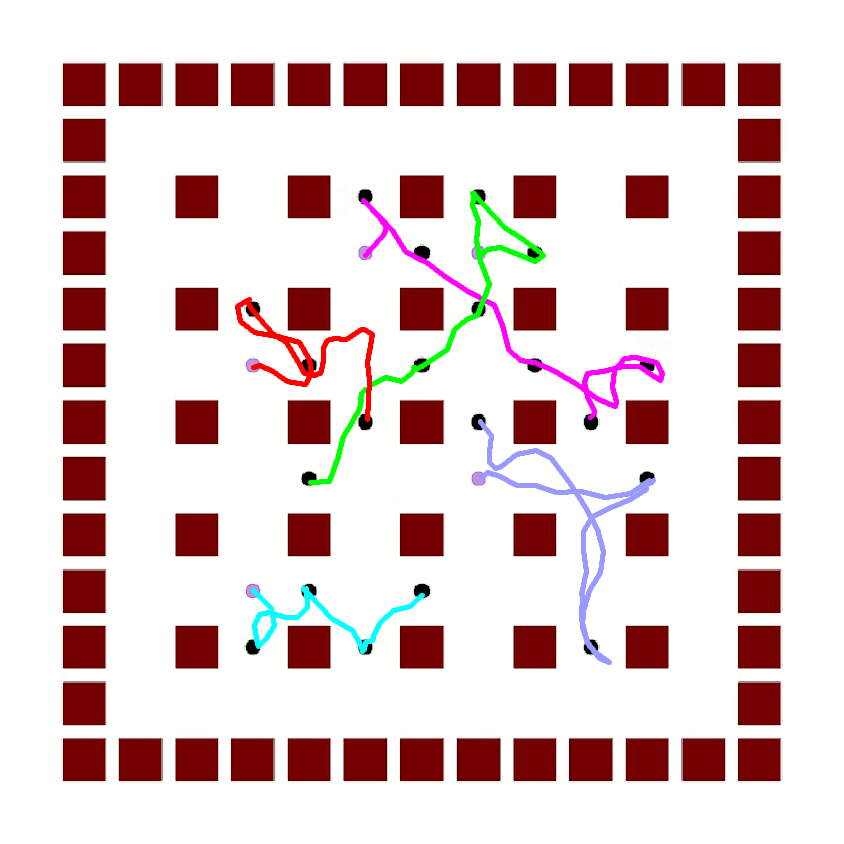} 
    \end{minipage}}

    \subfloat[]{  
    \begin{minipage}[b]{0.24\textwidth} 
        \centering
        \includegraphics[width=4cm,height=4cm]{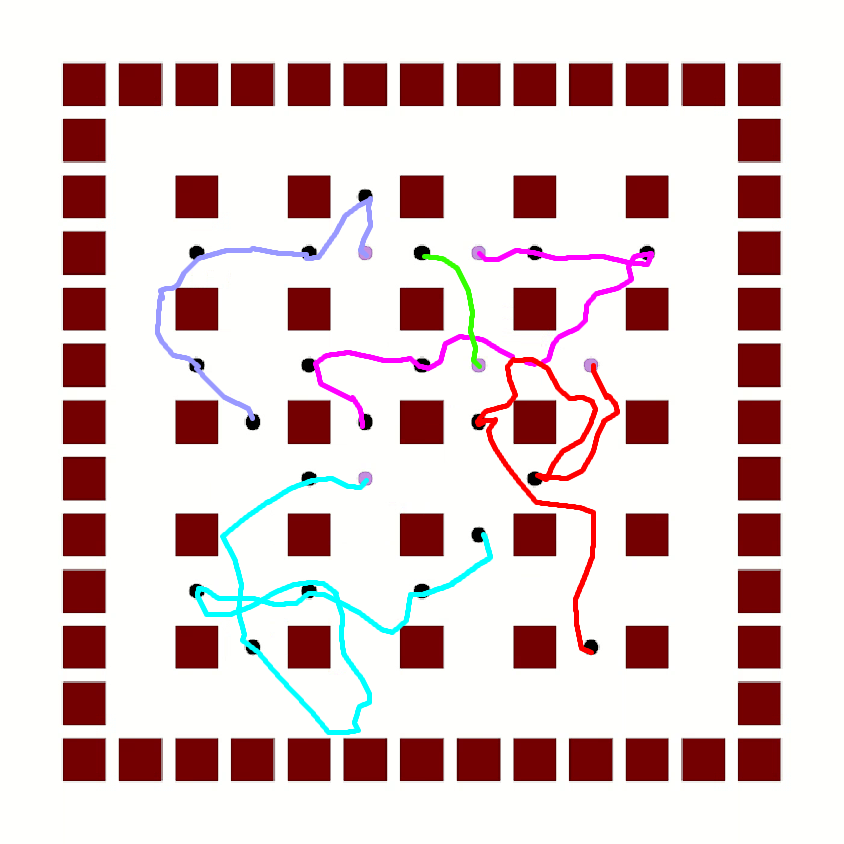} 
    \end{minipage}}
    \subfloat[]{ 
    \begin{minipage}[b]{0.24\textwidth} 
        \centering
        \includegraphics[width=4cm,height=4cm]{figures/a5_l20/a5_l20_r3_4.png} 
    \end{minipage}}
    \caption{Five agents - twenty tasks}
    \label{Figures 10} 
\end{figure}

Fig. \ref{Figures 5} shows the training average return curves. In all the different level scenarios, the average returns increase monotonously. This verifies the stability of the proposed method. Fig. \ref{Figures 6} to Fig. \ref{Figures 10} show the performance target assignment and pathfinding in five different level scenarios. Experimental results show that although the difficulty of these five different level scenarios increases gradually, the proposed method performs well in all the tasks. 

For easy tasks (\textit{two agents - two tasks}) in Fig. \ref{Figures 6} (a) (b) (c) (d), we can see that the target assignment and pathfinding were addressed well. For target assignment, results show that the task assignment is addressed very reasonably because the proposed method assigns the task to the agent that is close to the task. For pathfinding, it is shown that the planned paths are almost shortest. 

For difficult tasks (\textit{five agents - twenty tasks}) in Fig. \ref{Figures 10} (a) (b) (c) (d), we can see that the target assignment and pathfinding were also addressed well. For target assignment, it can be seen that each agent is reasonably assigned to several nearby tasks. For pathfinding, it is shown that the planned paths are usually the shortest. The same superior performances can be seen in other different level scenarios, such as \textit{two agents - four tasks} in Fig. \ref{Figures 7}, \textit{five agents - five tasks} in Fig. \ref{Figures 8}, and \textit{five agents - ten tasks} in Fig. \ref{Figures 9}. This demonstrated the efficiency of the proposed method.

\subsection{Cooperation Ability}
\label{Cooperation ability}
\begin{figure}[htbp]
	\centering
	\includegraphics[width=6cm,height=4.5cm]{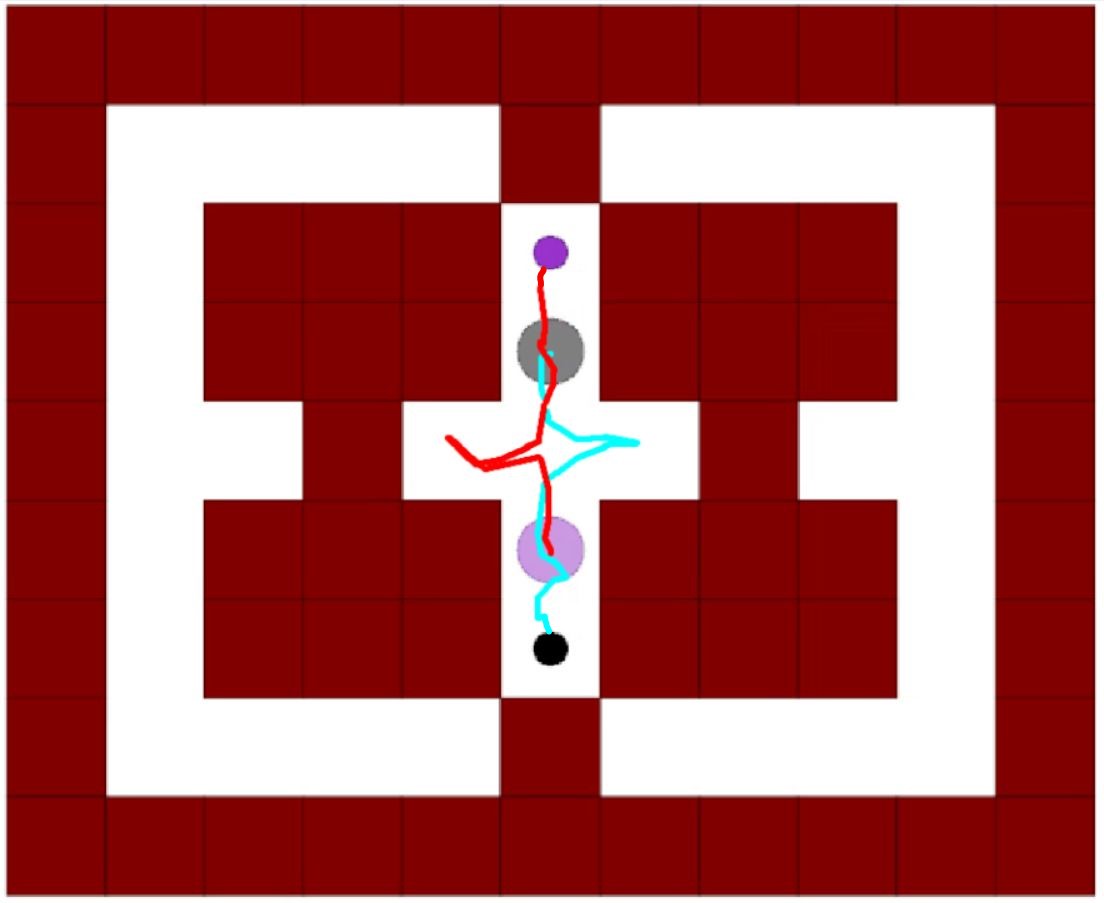}
	\caption{The cooperation of agents} 
	\label{Figures 11}
\end{figure}

We also designed a conflict scenario to verify the agents' learned cooperation ability. As shown in Fig. \ref{Figures 11}, the big purple circle represents agent-1, its task is denoted by the small purple circle. The big gray circle represents agent-2, and the small black circle denotes its task. We deliberately block other roads to create a conflict environment. As shown in Fig. \ref{Figures 11}, there is bound to be a conflict between agent-1 and agent-2 during the navigation. The trajectories generated by the proposed method are shown in Fig. \ref{Figures 11}. The red curve is the trajectory of agent-1, and the cyan curve is the trajectory of agent-2. Results show that both agent-1 and agent-2 learned to avoid each other at the point of conflict and then navigated to their tasks. This verified the cooperation ability of the proposed method.

\begin{table*}[ht]
    \begin{center}
        \begin{minipage}{\textwidth}
            \caption{Time comparisons between our method and traditional methods}
            \label{tab1}
            \begin{tabular*}{\textwidth}{@{\extracolsep{\fill}}lll@{\extracolsep{\fill}}}
                \toprule%
                Scenarios & Ours  & Traditional methods\\
                \midrule
                two agents - two tasks    & 0.0003221s(TAPF-ES)   & 0.02091s(TA)+0.0038107s(PF-ES)\\
                two agents - four tasks    & 0.0003287s(TAPF-ES)   & 0.03887s(TA)+0.0036540s(PF-ES)\\
                five agents - five tasks    & 0.0003375s(TAPF-ES)   & 0.0498669s(TA)+0.0037150s(PF-ES)\\
                five agents - ten tasks    & 0.0003414s(TAPF-ES)   & 0.2918257s(TA)+0.0054604s(PF-ES)\\
                five agents - twenty tasks    & 0.0003505s(TAPF-ES)   & 14.66s(TA)+0.0038555s(PF-ES)\\
                % \hline
                \bottomrule
            \end{tabular*}
        \end{minipage}
    \end{center}
\end{table*}

\subsection{Time Efficiency}
\label{Time efficiency}
Table \ref{tab1} shows the time comparisons between traditional methods \cite{gao2022two,xu2021multi} and the proposed method. Table \ref{tab1} shows that the proposed method simultaneously addresses target assignment (TA) and path planning (PF) problems. However, the traditional method addresses target assignment first and then performs path planning. Thus, the time consumed by the proposed method is TAPF time. However, the time consumed by the traditional method is TA+PF. In Table \ref{tab1}, ES means \textbf{e}ach \textbf{s}tep. The reasons that we compare the consumed time of each step are: (1) Our method handles continuous action space problems. However, the traditional method handles discrete action space problems. (2) The consumed time of path finding can be influenced by many factors, such as the resolution of the grid map and the distance taken at each step in the grid map. (3) Previous literature \cite{gao2022two,xu2021multi} did not consider the physical dynamics of agents. However, the physical dynamics of agents have been considered in this study. Thus, for a fair comparison, we only compare the consumed time of each step, which means the consumed time from an agent obtains an observation to output the decision action. For the traditional method, the consumed time of target assignment is computed by the method in \cite{gao2022two}, and the consumed time of each step in path planning is computed by the method in \cite{xu2021multi}.

Table \ref{tab1} shows that in easy tasks such \textit{two agents - two tasks} and \textit{two agents - four tasks}, the consumed time of target assignment in the traditional method can be acceptable. However, as the difficulty of tasks increased, the time of solving target assignments in the traditional method increased rapidly, especially in the \textit{five agents - twenty tasks} scenario. However, the proposed method provides policy time-efficiently in all different level tasks (from \textit{two agents - two tasks} to \textit{five agents - twenty tasks}). This verified the time efficiency ability of the proposed method. As we all know, time efficiency is an important factor in real-world TAPF problems. Therefore, the experiment results demonstrated the promising engineering practical value of the proposed method. Note that Fig. \ref{Figures 6} to Fig. \ref{Figures 10} had shown the performances of target assignment and path finding of the proposed method in five different level scenarios. For target assignment, results show that the task assignment is addressed reasonably because the proposed method assigns the task to the agent that is close to the task. For pathfinding, results show that the planned paths are almost shortest.

\section{Conclusion and future work}
\label{Conclusions}
This paper proposed a method to simultaneously solve the TAPF problem in the intelligent warehouse from a cooperative multi-agent deep RL perspective. First, we modeled the TAPF problem as a cooperative multi-agent deep RL problem. Then, we simultaneously solved task assignment and path finding using a cooperative multi-agent deep RL algorithm. Moreover, previous literature rarely considers the physical dynamics of agents. In this study, the physical dynamics of agents have been considered. Experimental results showed that the proposed method performed well in various task settings, which meant the target assignment was solved reasonably, and the planned path was almost the shortest. Furthermore, the proposed method is more time-efficient than baselines. For future work, we will apply the proposed method to real-world TAPF problems.

\bibliographystyle{IEEEtran}
\bibliography{ref}

% Generated by IEEEtran.bst, version: 1.14 (2015/08/26)
\begin{thebibliography}{10}
\providecommand{\url}[1]{#1}
\csname url@samestyle\endcsname
\providecommand{\newblock}{\relax}
\providecommand{\bibinfo}[2]{#2}
\providecommand{\BIBentrySTDinterwordspacing}{\spaceskip=0pt\relax}
\providecommand{\BIBentryALTinterwordstretchfactor}{4}
\providecommand{\BIBentryALTinterwordspacing}{\spaceskip=\fontdimen2\font plus
\BIBentryALTinterwordstretchfactor\fontdimen3\font minus \fontdimen4\font\relax}
\providecommand{\BIBforeignlanguage}[2]{{%
\expandafter\ifx\csname l@#1\endcsname\relax
\typeout{** WARNING: IEEEtran.bst: No hyphenation pattern has been}%
\typeout{** loaded for the language `#1'. Using the pattern for}%
\typeout{** the default language instead.}%
\else
\language=\csname l@#1\endcsname
\fi
#2}}
\providecommand{\BIBdecl}{\relax}
\BIBdecl

\bibitem{adhau2012multi}
S.~Adhau, M.~L. Mittal, and A.~Mittal, ``A multi-agent system for distributed multi-project scheduling: An auction-based negotiation approach,'' \emph{Engineering Applications of Artificial Intelligence}, vol.~25, no.~8, pp. 1738--1751, 2012.

\bibitem{ayari2019acd3gpso}
A.~Ayari and S.~Bouamama, ``Acd3gpso: automatic clustering-based algorithm for multi-robot task allocation using dynamic distributed double-guided particle swarm optimization,'' \emph{Assembly Automation}, vol.~40, no.~2, pp. 235--247, 2020.

\bibitem{wurman2008coordinating}
P.~R. Wurman, R.~D'Andrea, and M.~Mountz, ``Coordinating hundreds of cooperative, autonomous vehicles in warehouses,'' \emph{AI Magazine}, vol.~29, no.~1, pp. 9--9, 2008.

\bibitem{wan2018lifelong}
Q.~Wan, C.~Gu, S.~Sun, M.~Chen, H.~Huang, and X.~Jia, ``Lifelong multi-agent path finding in a dynamic environment,'' in \emph{2018 15th International Conference on Control, Automation, Robotics and Vision (ICARCV)}.\hskip 1em plus 0.5em minus 0.4em\relax IEEE, 2018, pp. 875--882.

\bibitem{ma2016multi}
H.~Ma, C.~Tovey, G.~Sharon, T.~S. Kumar, and S.~Koenig, ``Multi-agent path finding with payload transfers and the package-exchange robot-routing problem,'' in \emph{Proceedings of the Thirtieth AAAI Conference on Artificial Intelligence}, vol.~30, no.~1, 2016, pp. 3166--3173.

\bibitem{silver2014deterministic}
D.~Silver, G.~Lever, N.~Heess, T.~Degris, D.~Wierstra, and M.~Riedmiller, ``Deterministic policy gradient algorithms,'' in \emph{International Conference on Machine Learning (ICML)}.\hskip 1em plus 0.5em minus 0.4em\relax PMLR, 2014, pp. 387--395.

\bibitem{10476692}
P.~Ladosz, M.~Mammadov, H.~Shin, W.~Shin, and H.~Oh, ``Autonomous landing on a moving platform using vision-based deep reinforcement learning,'' \emph{IEEE Robotics and Automation Letters}, vol.~9, no.~5, pp. 4575--4582, 2024.

\bibitem{10508809}
Q.~Liu, Y.~Li, X.~Shi, K.~Lin, Y.~Liu, and Y.~Lou, ``Distributional policy gradient with distributional value function,'' \emph{IEEE Transactions on Neural Networks and Learning Systems}, pp. 1--13, 2024.

\bibitem{rashid2018qmix}
T.~Rashid, M.~Samvelyan, C.~Schroeder, G.~Farquhar, J.~Foerster, and S.~Whiteson, ``Qmix: Monotonic value function factorisation for deep multi-agent reinforcement learning,'' in \emph{International Conference on Machine Learning}.\hskip 1em plus 0.5em minus 0.4em\relax PMLR, 2018, pp. 4295--4304.

\bibitem{10466624}
Q.~Liu, Y.~Li, Y.~Liu, K.~Lin, J.~Gao, and Y.~Lou, ``Data efficient deep reinforcement learning with action-ranked temporal difference learning,'' \emph{IEEE Transactions on Emerging Topics in Computational Intelligence}, vol.~8, no.~4, pp. 2949--2961, 2024.

\bibitem{gerkey2004formal}
B.~P. Gerkey and M.~J. Matari{\'c}, ``A formal analysis and taxonomy of task allocation in multi-robot systems,'' \emph{The International Journal of Robotics Research}, vol.~23, no.~9, pp. 939--954, 2004.

\bibitem{liu2012centralized}
C.~Liu and A.~Kroll, ``A centralized multi-robot task allocation for industrial plant inspection by using {A}* and genetic algorithms,'' in \emph{International Conference on Artificial Intelligence and Soft Computing}.\hskip 1em plus 0.5em minus 0.4em\relax Springer, 2012, pp. 466--474.

\bibitem{glover1990tabu}
F.~Glover, ``Tabu search: A tutorial,'' \emph{Interfaces}, vol.~20, no.~4, pp. 74--94, 1990.

\bibitem{kuhn1955hungarian}
H.~W. Kuhn, ``The hungarian method for the assignment problem,'' \emph{Naval Research Logistics Quarterly}, vol.~2, no. 1-2, pp. 83--97, 1955.

\bibitem{giordani2013distributed}
S.~Giordani, M.~Lujak, and F.~Martinelli, ``A distributed multi-agent production planning and scheduling framework for mobile robots,'' \emph{Computers \& Industrial Engineering}, vol.~64, no.~1, pp. 19--30, 2013.

\bibitem{best2019dec}
G.~Best, O.~M. Cliff, T.~Patten, R.~R. Mettu, and R.~Fitch, ``Dec-mcts: Decentralized planning for multi-robot active perception,'' \emph{The International Journal of Robotics Research}, vol.~38, no. 2-3, pp. 316--337, 2019.

\bibitem{bernstein2002complexity}
D.~S. Bernstein, R.~Givan, N.~Immerman, and S.~Zilberstein, ``The complexity of decentralized control of markov decision processes,'' \emph{Mathematics of Operations Research}, vol.~27, no.~4, pp. 819--840, 2002.

\bibitem{zavlanos2008distributed}
M.~M. Zavlanos, L.~Spesivtsev, and G.~J. Pappas, ``A distributed auction algorithm for the assignment problem,'' in \emph{2008 47th IEEE Conference on Decision and Control}.\hskip 1em plus 0.5em minus 0.4em\relax IEEE, 2008, pp. 1212--1217.

\bibitem{wagner2015subdimensional}
G.~Wagner and H.~Choset, ``Subdimensional expansion for multirobot path planning,'' \emph{Artificial Intelligence}, vol. 219, pp. 1--24, 2015.

\bibitem{sharon2015conflict}
G.~Sharon, R.~Stern, A.~Felner, and N.~R. Sturtevant, ``Conflict-based search for optimal multi-agent pathfinding,'' \emph{Artificial Intelligence}, vol. 219, pp. 40--66, 2015.

\bibitem{sharon2013increasing}
G.~Sharon, R.~Stern, M.~Goldenberg, and A.~Felner, ``The increasing cost tree search for optimal multi-agent pathfinding,'' \emph{Artificial Intelligence}, vol. 195, pp. 470--495, 2013.

\bibitem{nguyen2019generalized}
V.~Nguyen, P.~Obermeier, T.~Son, T.~Schaub, and W.~Yeoh, ``Generalized target assignment and path finding using answer set programming,'' in \emph{Proceedings of the International Symposium on Combinatorial Search}, vol.~10, no.~1, 2019, pp. 194--195.

\bibitem{silver2005cooperative}
D.~Silver, ``Cooperative pathfinding,'' in \emph{Proceedings of the AAAI conference on Artificial Intelligence and Interactive Digital Entertainment}, vol.~1, 2005, pp. 117--122.

\bibitem{surynek2009novel}
P.~Surynek, ``A novel approach to path planning for multiple robots in bi-connected graphs,'' in \emph{2009 IEEE International Conference on Robotics and Automation}.\hskip 1em plus 0.5em minus 0.4em\relax IEEE, 2009, pp. 3613--3619.

\bibitem{surynek2017modifying}
P.~Surynek, A.~Felner, R.~Stern, and E.~Boyarski, ``Modifying optimal sat-based approach to multi-agent path-finding problem to suboptimal variants,'' in \emph{International Symposium on Combinatorial Search}, vol.~8, 2017, pp. 169--170.

\bibitem{barer2014suboptimal}
M.~Barer, G.~Sharon, R.~Stern, and A.~Felner, ``Suboptimal variants of the conflict-based search algorithm for the multi-agent pathfinding problem,'' in \emph{Proceedings of the International Symposium on Combinatorial Search}, vol.~5, no.~1, 2014, pp. 19--27.

\bibitem{peihuang2009path}
L.~Peihuang, W.~Xing, and W.~Jiarong, ``Path planning and control for multiple agvs based on improved two-stage traffic scheduling,'' \emph{International Journal of Automation Technology}, vol.~3, no.~2, pp. 157--164, 2009.

\bibitem{sutton2018reinforcement}
R.~S. Sutton and A.~G. Barto, \emph{Reinforcement learning: An introduction}.\hskip 1em plus 0.5em minus 0.4em\relax MIT press, 2018.

\bibitem{sartoretti2019primal}
G.~Sartoretti, J.~Kerr, Y.~Shi, G.~Wagner, T.~S. Kumar, S.~Koenig, and H.~Choset, ``Primal: Pathfinding via reinforcement and imitation multi-agent learning,'' \emph{IEEE Robotics and Automation Letters}, vol.~4, no.~3, pp. 2378--2385, 2019.

\bibitem{tampuu2017multiagent}
A.~Tampuu, T.~Matiisen, D.~Kodelja, I.~Kuzovkin, K.~Korjus, J.~Aru, J.~Aru, and R.~Vicente, ``Multiagent cooperation and competition with deep reinforcement learning,'' \emph{PloS one}, vol.~12, no.~4, p. e0172395, 2017.

\bibitem{hernandez2017survey}
P.~Hernandez-Leal, M.~Kaisers, T.~Baarslag, and E.~M. de~Cote, ``A survey of learning in multiagent environments: Dealing with non-stationarity,'' \emph{arXiv preprint arXiv:1707.09183}, 2017.

\bibitem{sunehag2018value}
P.~Sunehag, G.~Lever, A.~Gruslys, W.~M. Czarnecki, V.~Zambaldi, M.~Jaderberg, M.~Lanctot, N.~Sonnerat, J.~Z. Leibo, K.~Tuyls \emph{et~al.}, ``Value-decomposition networks for cooperative multi-agent learning based on team reward,'' in \emph{Proceedings of the 17th International Conference on Autonomous Agents and MultiAgent Systems}, 2018, pp. 2085--2087.

\bibitem{lowe2017multi}
R.~Lowe, Y.~Wu, A.~Tamar, J.~Harb, P.~Abbeel, and I.~Mordatch, ``Multi-agent actor-critic for mixed cooperative-competitive environments,'' in \emph{Proceedings of the 31st International Conference on Neural Information Processing Systems}, vol.~30, 2017, pp. 6382--6393.

\bibitem{kraemer2016multi}
L.~Kraemer and B.~Banerjee, ``Multi-agent reinforcement learning as a rehearsal for decentralized planning,'' \emph{Neurocomputing}, vol. 190, no.~C, pp. 82--94, 2016.

\bibitem{9366340}
M.~Damani, Z.~Luo, E.~Wenzel, and G.~Sartoretti, ``Primal2: Pathfinding via reinforcement and imitation multi-agent learning - lifelong,'' \emph{IEEE Robotics and Automation Letters}, vol.~6, no.~2, pp. 2666--2673, 2021.

\bibitem{ROSplanning}
D.~H. Michael~Ferguson, David V.~Lu, ``Ros-planning/navigation,'' \url{https://github.com/ros-planning/navigation}, 2022.

\bibitem{gao2022two}
J.~Gao, Y.~Li, Y.~Xu, and S.~Lv, ``A two-objective ilp model of op-matsp for the multi-robot task assignment in an intelligent warehouse,'' \emph{Applied Sciences}, vol.~12, no.~10, p. 4843, 2022.

\bibitem{xu2021multi}
Y.~Xu, Y.~Li, Q.~Liu, J.~Gao, Y.~Liu, and M.~Chen, ``Multi-agent pathfinding with local and global guidance,'' in \emph{2021 IEEE International Conference on Networking, Sensing and Control (ICNSC)}, vol.~1, 2021, pp. 1--7.

\end{thebibliography}

\end{document}